\documentclass{article}

\usepackage[preprint]{corl_2025} % Uncomment for pre-prints (e.g., arxiv); This is like ``final'', but will remove the CORL footnote.
\usepackage{times}

\usepackage[numbers]{natbib}
\usepackage{multicol}

\usepackage{graphicx}
\usepackage{amsmath}
\usepackage{amssymb}
\usepackage{bm}
\usepackage{booktabs}
\usepackage{comment}
\usepackage[capitalize]{cleveref}
\crefname{section}{Sec.}{Secs.}
\Crefname{section}{Section}{Sections}
\Crefname{table}{Table}{Tables}
\crefname{table}{Tab.}{Tabs.}

\usepackage[toc]{appendix}
\usepackage{etoc}
\usepackage{minitoc}
\etocsettocstyle{\section*{Appendix}}{}

\newcommand{\method}{3DFA}
\newcommand{\methodfull}{3D FlowMatch Actor}

\newcommand{\model}{\text{3D FlowMatch Actor}}

\usepackage{amsmath,amsfonts,bm}

\def\imw#1#2{\includegraphics[width=#2\linewidth]{#1.png}}
\def\imwjpg#1#2{\includegraphics[width=#2\linewidth]{#1.jpg}}

\newcommand{\tb}[3]{\setlength{\tabcolsep}{#2mm}\begin{tabular}{#1}#3\end{tabular}}

\newcommand{\xmark}{\textcolor{red}{\ding{55}}}

\def\eqref#1{equation~\ref{#1}}
\def\1{\bm{1}}

\DeclareMathAlphabet{\mathsfit}{\encodingdefault}{\sfdefault}{m}{sl}
\SetMathAlphabet{\mathsfit}{bold}{\encodingdefault}{\sfdefault}{bx}{n}

\def\gN{{\mathcal{N}}}

\newcommand{\ac}{\mathbf{a}}
\newcommand{\oo}{\mathbf{o}}

\newcommand{\lang}{l}
\newcommand{\hist}{c}

\newcommand{\id}{i}

\newcommand{\tr}{\bm{\tau}}
\usepackage{graphicx}
\usepackage{amsmath}
\usepackage{amssymb}
\usepackage{booktabs}

\usepackage[utf8]{inputenc} % allow utf-8 input
\usepackage[T1]{fontenc}    % use 8-bit T1 fonts
\usepackage{hyperref}       % hyperlinks
\usepackage{url}            % simple URL typesetting
\usepackage{amsfonts}       % blackboard math symbols
\usepackage{nicefrac}       % compact symbols for 1/2, etc.
\usepackage{microtype}      % microtypography

\usepackage{listings}
\usepackage{mwe} % for placeholder images
\usepackage{makecell}
\usepackage{color, colortbl}
\usepackage[normalem]{ulem} % strikethrough font
\usepackage[ruled,vlined]{algorithm2e}
\usepackage{algorithmic}

\usepackage{wrapfig}
\usepackage{caption}
\usepackage{pifont} % Checkmark and Crossmark
\usepackage{multirow}
\usepackage{float}

\usepackage{chngpage}
\usepackage{wrapfig}
\usepackage{adjustbox}
\usepackage{comment}
\usepackage{color}
\usepackage{xcolor}

\usepackage{microtype} 
\usepackage{listings}

\definecolor{Gray}{gray}{0.9}
\definecolor{ImportantColor}{rgb}{0.63, 0.79, 0.95}
\newcolumntype{g}{>{\columncolor{ImportantColor}}c}
\newcolumntype{?}{!{\vrule width 1pt}}
\newcommand{\cmark}{\ding{51}}%

\def\imw#1#2{\includegraphics[width=#2\linewidth]{#1.png}}
\def\imwjpg#1#2{\includegraphics[width=#2\linewidth]{#1.jpg}}

\graphicspath{
{figs/},
}

\newcommand*{\belowrulesepcolor}[1]{% 
  \noalign{% 
    \kern-\belowrulesep 
    \begingroup 
      \color{#1}% 
      \hrule height\belowrulesep 
    \endgroup 
  }%
} 
\newcommand*{\aboverulesepcolor}[1]{% 
  \noalign{% 
    \begingroup 
      \color{#1}% 
      \hrule height\aboverulesep 
    \endgroup 
    \kern-\aboverulesep 
  }%
}

\definecolor{codegreen}{rgb}{0,0.6,0}
\definecolor{codegray}{rgb}{0.5,0.5,0.5}
\definecolor{codepurple}{rgb}{0.58,0,0.82}
\definecolor{backcolour}{rgb}{0.95,0.95,0.92}

\lstdefinestyle{mystyle}{
    backgroundcolor=\color{backcolour},   
    commentstyle=\color{codegreen},
    keywordstyle=\color{magenta},
    numberstyle=\tiny\color{codegray},
    stringstyle=\color{codepurple},
    basicstyle=\ttfamily\footnotesize,
    breakatwhitespace=false,         
    breaklines=true,                 
    captionpos=b,                    
    keepspaces=true,                 
    numbers=left,                    
    numbersep=5pt,                  
    showspaces=false,                
    showstringspaces=false,
    showtabs=false,                  
    tabsize=2
}

\title{\methodfull{}: An Efficient    3D  Policy for Robot  Manipulation}
\title{\methodfull{}: Unified 3D Policy for Single- and Dual-Arm Manipulation}

\author{
  \textbf{Nikolaos Gkanatsios}$^{\dagger,1}$ \quad
  \textbf{Jiahe Xu}$^{\dagger,1}$ \quad
  \textbf{Matthew Bronars}$^{1}$ \quad
  \textbf{Arsalan Mousavian}$^{2}$ \\
  \textbf{Tsung-Wei Ke}$^{\star,3}$ \quad
  \textbf{Katerina Fragkiadaki}$^{1}$ \\
  \normalfont
  $^{1}$Carnegie Mellon University \quad
  $^{2}$NVIDIA \quad
  $^{3}$National Taiwan University \\
  \textcolor{magenta}{\url{https://3d-flowmatch-actor.github.io/}}
}

\begin{document}

\maketitle
\let\thefootnote\relax\footnotetext{$^{\dagger}$Equal contribution, $^{\star}$Work done at CMU}

\etocdepthtag.toc{mtchapter}
\etocsettagdepth{mtchapter}{subsection}
\etocsettagdepth{mtappendix}{none}
\faketableofcontents

\begin{abstract} 
We present \methodfull{} (\method{}), a 3D policy architecture for robot manipulation that combines flow matching for trajectory prediction with 3D pretrained visual scene representations for learning from demonstration. \method{} leverages 3D relative attention between action and visual tokens during action denoising, building on prior work in 3D diffusion-based single-arm policy learning. Through a combination of flow matching and targeted system-level and architectural optimizations, \method{} achieves over 30× faster training and inference than previous 3D diffusion-based policies, without sacrificing performance. On the bimanual PerAct2 benchmark, it establishes a new state of the art, outperforming the next-best method by an absolute margin of 41.4\%. In extensive real-world evaluations, it surpasses strong baselines with up to 1000× more parameters and significantly more pretraining. In unimanual settings, it sets a new state of the art on 74 RLBench tasks by directly predicting dense end-effector trajectories, eliminating the need for motion planning. Comprehensive ablation studies underscore the importance of our design choices for both policy effectiveness and efficiency.

\end{abstract} 

% \keywords{Bimanual manipulation, Single-arm manipulation, Flow matching, 3D Policy, Imitation learning}
\section{Introduction}
\label{sec:intro}

Single-arm manipulation has achieved great success in handling long-horizon and high-precision tasks~\cite{liu2024libero,james2020rlbench,li2024evaluating}, even in highly cluttered environments~\cite{khazatsky2024droid,fang2023rh20t}.  However, the lack of coordination between multiple end-effectors largely constrains single-arm systems to simple pick-and-place tasks, making them inadequate for addressing the more complex and diverse manipulation challenges encountered in real-world daily tasks. To overcome these challenges, bimanual systems offer a promising alternative by enabling more dexterous and coordinated interactions with the environment~\cite{smith2012dual}.

Although bimanual setups improve the ability of a robot to perform more intricate and dexterous tasks, they also impose stricter demands on spatiotemporal precision. Both arms must operate in a tightly coordinated manner, executing actions in the correct temporal sequence and at precisely aligned spatial locations. This added complexity makes learning effective manipulation policies more difficult than in the single-arm case. Despite growing interest, existing approaches~\cite{Grannen2022LearningBS,zhao2023learningfinegrainedbimanualmanipulation,grannen2023stabilizeactlearningcoordinate,grotz2024peract2,Liu2024VoxActBVA} still fall short of achieving robust generalization across a wide range of tasks.

In parallel, recent advances in single-arm manipulation have demonstrated the power of diffusion models in capturing multimodal behaviors~\cite{ke20243d,chi2023diffusion,reuss2023goal,Ze2024DP3}, achieving high-precision action prediction through 3D scene understanding~\cite{shridhar2023perceiver,gervet2023act3d,goyal2023rvt,xian2023chaineddiffuser} and impressive generalization to various tasks and language instructions~\cite{black2024pi_0,li2024cogact,qu2025spatialvla}. A natural next step toward robust bimanual manipulation is to integrate these advances. In fact, we show that an adaptation of 3D Diffuser Actor (3DDA)~\cite{ke20243d} - a model that combines diffusion-based action generation with 3D scene representations - already establishes a new state of the art on the bimanual manipulation benchmark PerAct2~\cite{grotz2024peract2}. 
Next, we ask the question: what prevents 3DDA from being deployed in real-world bimanual manipulation scenarios? Our experimentation suggests two key bottlenecks: slow inference speed and long training time. For instance, on PerAct2, 3DDA requires approximately 21 days to train and operates at 0.5Hz during inference. The prolonged training time significantly limits the model's ability to adapt to new tasks, while the slow inference speed makes it unsuitable for real-time or dynamic task execution.

In light of these observations, we introduce \methodfull{} (\method{}), a significantly more efficient extension of 3DDA that improves both training time and inference speed by an order of magnitude. On the PerAct2 bimanual manipulation benchmark~\cite{grotz2024peract2}, \method{} reduces the training time from 21 days to 16 hours and increases the inference speed from 0.5Hz to 18.2Hz, without sacrificing performance. 
To enable faster inference, we replace the DDPM-based~\cite{DDPM} formulation used in 3DDA with a Flow Matching approach~\cite{liu2022flow,lipman2022flow}, reducing the number of denoising steps during inference from 100 to 5. 
To reduce training overhead, we implement a series of system-level optimizations for computational efficiency, such as faster token sampling, fewer camera inputs, optimized data loading, efficient attention implementation and mixed-precision training.
While none of these techniques is novel on its own, our contribution lies in carefully integrating them into a manipulation system. 

\method{} achieves a new state of the art on the PerAct2 simulation benchmark, with a success rate of $85.1\%$. Furthermore, it outperforms strong baselines—including $\pi_0$~\cite{black2024pi_0}—both in simulation and on a real-world 10-task benchmark we constructed using the bimanual ALOHA platform~\cite{fu2024mobilealohalearningbimanual}. We conduct an extensive ablation study to break down the contributions of each design choice.

Notably, \method{} is a general-purpose framework capable of predicting both sparse keyposes and dense end-effector trajectories, and is applicable to both unimanual and bimanual manipulation. On the 18 single-arm tasks of PerAct~\cite{shridhar2023perceiver}, \method{} is trained to predict the next end-effector keypose and performs on par with 3DDA, while requiring significantly less training and inference time. Furthermore, on the 74-task benchmark of~\cite{hiveformer}, \method{} excels at jointly predicting the next keypose and the trajectory connecting it to the current pose in a single forward pass, outperforming the best baseline by $7.3\%$. These results highlight the versatility and efficiency of \method{} as a framework for 3D manipulation.

In summary, our contributions are: (1) Adaptation of state-of-the-art single-arm 3D generative policies to bimanual manipulation, (2) Dramatic acceleration of training and inference time in 3D policies, (3) State-of-the-art bimanual manipulation results on PerAct2, with an absolute margin of 41.4\% over $\pi_0$, (4) State-of-the-art bimanual manipulation results in the real world, outperforming foundational policies in a direct comparison, (5) State of the art on the HiveFormer unimanual 74-task benchmark, demonstrating planner-free trajectory prediction capabilities.

% \input{arxiv_3_relatedwork}
%%%%%%%%%%%%%%%%%%%%%%%%%%%%%%%%%%%%%%%%%%%%%%%%%%%
\begin{figure*}[t]
    % \vspace{-20pt}
    \centering
    \begin{adjustbox}{center}
     \includegraphics[width=1.0\textwidth,keepaspectratio]{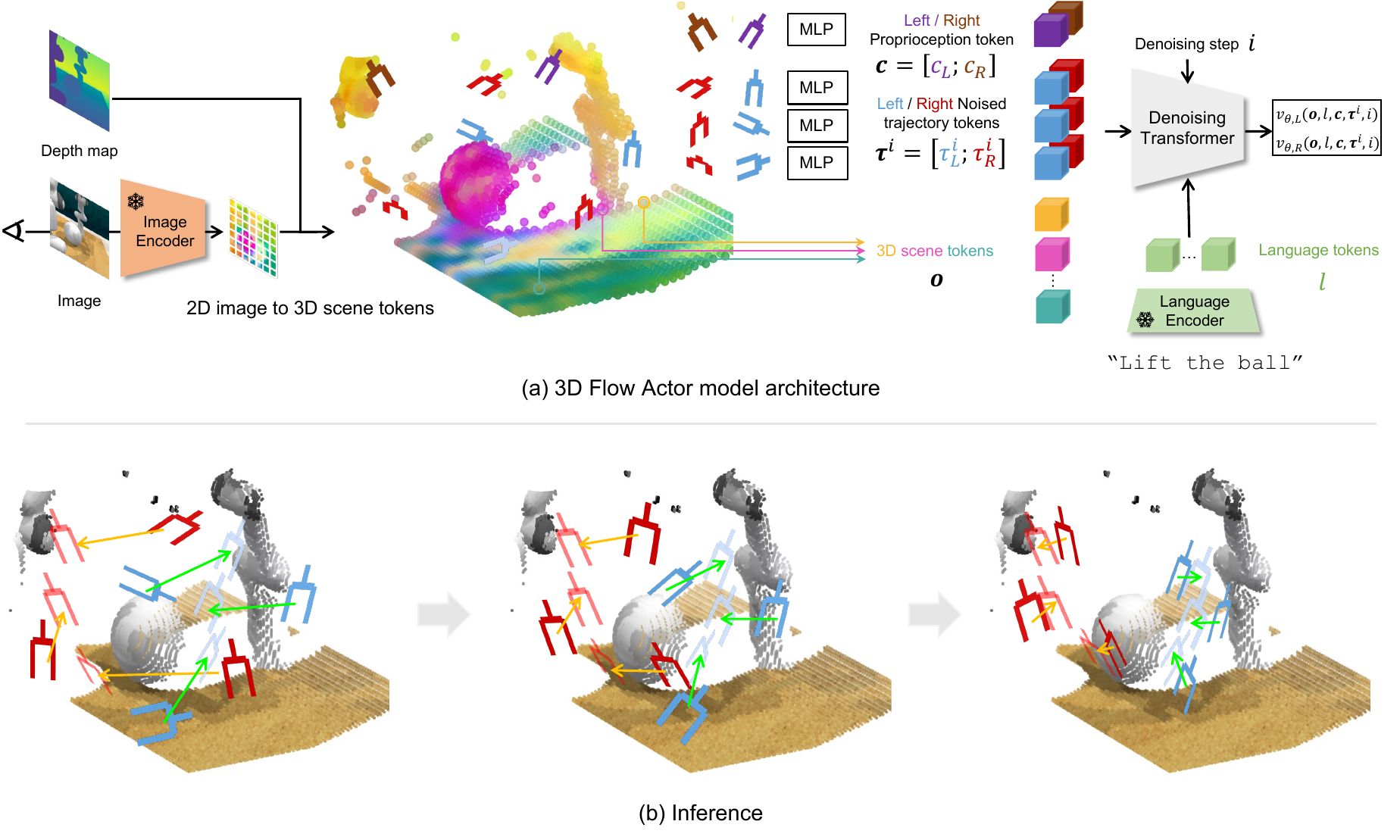}
    \end{adjustbox}    
    \caption{\textbf{Top:} \method{} is a flow-matching policy built atop 3D Diffuser Actor~\cite{ke20243d}. It encodes the visual scene $\oo$, robot proprioception for left and right arms $\hist_L$, $\hist_R$, and noised trajectories $\tr^\id_L$, $\tr^\id_R$ into 3D tokens. Given language tokens $l$, diffusion step $i$, and these 3D tokens, \method{} predicts velocity fields $v_{\theta,L}$ and $v_{\theta,R}$ for the left and right arms, respectively. \textbf{Bottom:} During inference, \method{} iteratively predicts straight-line velocity fields pointing toward the {\color{green}left-} and {\color{orange}right-}hand target poses.
    }
    \label{fig:architecture}
    \vspace{-10pt}
\end{figure*}

\section{\model{}}
\label{sec:method}

The architecture of \method{} is shown in Figure~\ref{fig:architecture}. 
It is a robot policy for single- and dual-arm manipulation that generates 3D end-effector trajectories for one or more robot arms conditioned on the task instruction, proprioception history and scene visual information, through iterative denoising.  
\method{}  grounds visual and action tokens to 3D locations described in a common coordinate frame, using calibration information to transfer visual tokens from the camera frame to the robot's frame.

\method{} builds upon the state-of-the-art single-arm 3D diffusion policy, 3DDA~\cite{ke20243d}, which we extend to also solve bimanual tasks. To accelerate inference and training, we replace the original DDPM-based diffusion mechanism with flow matching, adopt faster point sampling methods and attention implementations, and build a highly-efficient data loading pipeline. % and refactor the model to support CUDA graph compilation.

We denote demonstrations as sequences of observations and actions $\{(\oo_1, \ac_1), (\oo_2, \ac_2), \ldots\}$, accompanied by a task language instruction $\lang$, where $\oo_t$ denotes the visual observation and $\ac_t$ the robot action at timestep $t$. Each observation $\oo_t$ consists of one or more posed RGB-D images. Each action $\ac_t$ is a single-arm end-effector pose and is decomposed into 3D location, rotation and binary (open/close) state: $\ac_t = \{ \ac_t^{\mathrm{\small loc}} \in \mathbb{R}^3, \ac_t^{\mathrm{\small rot}} \in \mathbb{R}^6, \ac_t^{\mathrm{\small open}} \in \{0,1\}\} $. Let $\tr_t=(\ac_{t:t+T}^{\mathrm{\small loc}}, \ac_{t:t+T}^{\mathrm{\small rot}})$ denote the trajectory of 3D locations and rotations at timestep $t$, with a temporal horizon $T$. At each timestep $t$, our model predicts one of more trajectories  $\tr_t$ and binary states $\ac_{t:t+T}^{\mathrm{\small open}}$. 
We first review Flow Matching in Section \ref{sec:flowmatch} and the 3DDA architecture in Section \ref{sec:DDA}. We then detail our extensions to support bimanual manipulation and the design choices to enhance efficiency in Section~\ref{sec:3dfa_core}.

\subsection{Flow Matching for Fast Action Generation}
\label{sec:flowmatch}
Diffusion models~\cite{DDPM,song2020score} are powerful generative frameworks for modeling multimodal data distributions. They generate samples by iteratively removing Gaussian noise via a stochastic process defined by conditional probability transitions. In contrast, flow matching approaches~\cite{lipman2022flow,liu2022flow} generate data by solving an optimal transport problem between a source $\mu_0$ and a target distribution $\mu_1$.

In particular, we adopt Rectified Flow~\cite{liu2022flow}, an instantiation of flow matching that transforms a sample $X_0 \sim \mu_0$ into $X_1 \sim \mu_1$ by following straight paths in the sample space. This significantly reduces computation during inference while retaining the expressiveness needed for high-dimensional generation, making it especially well-suited for robotics, where real-time performance is critical. 
The rectified flow between $(X_0, X_1)$ defines a continuous, time-differentiable trajectory $\mathbf{Z} = {Z_t : t \in [0, 1]}$ that transports $X_0$ to $X_1$ and is governed by the ordinary differential equation (ODE):
\begin{align}
    dZ_t = v^X_t(Z_t)dt, \quad t \in [0, 1], \quad Z_0 = X_0,
\end{align}
where $v^X_t$ is a time-dependent velocity field. The optimal velocity field that minimizes the expected discrepancy from the straight-line path between $X_0$ and $X_1$ can be formally defined as: $\inf_{v} \int^1_0 \mathbb{E}[\|X_1-X_0-v(X_t, t)\|]dt$, where $X_t = (1 - t)X_0 + tX_1$ denotes the linear interpolation at time $t$ \cite{liu2022flow}. 
The model is trained to approximate this velocity field by minimizing the loss \cite{liu2022flow}:
\begin{align}
    \mathcal{L}_{\theta}=\mathbb{E}_{t, X}[\|v_{\theta}(X_t, t) - v^X_t\|^2] 
\end{align}
In our setting, the goal is to generate robot actions from noise. We define $\mu_0 \sim \mathcal{N}(0, I)$ and $\mu_1$ as the distribution over real actions. During inference, the model iteratively transforms the noise sample $X_0$ into a predicted action $X_1$ over $N$ steps with a fixed step size $\Delta t = \frac{1}{N}$: $X_{t + \Delta t} = X_t - \Delta t \cdot v_{\theta}(X_t, t)$.

\subsection{3D Diffuser Actor}
\label{sec:DDA}
3DDA~\cite{ke20243d} is a 3D diffusion policy for manipulation that learns from demonstrations how to predict end-effector trajectories. By framing trajectory generation as a denoising process, 3DDA learns to iteratively refine noisy trajectory samples into clean end-effector motions, conditioning on multi-view RGB-D observations, language instructions and proprioception history. 3DDA conditions on 3D scene feature representations derived from posed cameras and sensed depth and uses DDPM-based diffusion to predict the noise component at each diffusion step. It distills 2D foundational features on point clouds to describe the scene and uses calibration information to transform the positional encodings of the  3D visual tokens into the robot's frame and to fuse them with action tokens that eventually decode the end-effector's translation and rotation at each future timestep. 

To simplify notation, we denote $\tr^\id$ as the noisy trajectory estimate at diffusion step $i$ without specifying the trajectory timestep.  The model conditions on the following inputs: (1) 3D scene tokens: 3DDA featurizes image views using a 2D image encoder and lifts each of the feature patches to 3D by calculating the average 3D location of each patch; (2) 3D proprioception tokens: 3DDA contextualizes a set of learnable embeddings with 3D scene tokens based on the proprioceptive location; (3) 3D trajectory tokens:  3DDA maps each noisy action $\ac^\id$ of trajectory $\tr^\id$ at diffusion step $i$ to a latent feature vector and lifts these feature vectors to 3D, based on the noisy location estimate of $\ac^\id$; (4) language tokens: language instructions are encoded to latent embeddings with a text encoder.  3DDA fuses all 3D tokens using relative 3D attentions, and additionally fuses language tokens using standard attention. As a last step, the refined trajectory tokens are passed through MLPs to predict the noise added to $\tr^0$, as well as the end-effector opening.  We refer to \cite{ke20243d} for more details.

\subsection{\methodfull{}}
\label{sec:3dfa_core}
We detail the core elements of \method{}, as changes we have made over the model of \cite{ke20243d}.

\paragraph{Extension to Bimanual Manipulation:}
We first redefine the robot action in a bimanual form: $\ac_{t, L}$ and $\ac_{t, R}$ denote the robot action at timestep $t$, of the left and right robot arm respectively.  Our goal is to predict the corresponding trajectory $\tr_{t, L}=(\ac_{t:t+T,L}^{\mathrm{\small loc}}, \ac_{t:t+T,L}^{\mathrm{\small rot}})$ and $\tr_{t,R}=(\ac_{t:t+T,R}^{\mathrm{\small loc}}, \ac_{t:t+T,r}^{\mathrm{\small rot}})$ of temporal horizon $T$ for both arms. To apply 3DFA on unimanual manipulation setups, we still use the unimanual trajectory definition: $\ac_{t}$ and $\tr_{t}=(\ac_{t:t+T}^{\mathrm{\small loc}}, \ac_{t:t+T}^{\mathrm{\small rot}})$. We follow the same 3D tokenization procedure to map (1) the noisy estimate of pose $\ac^i_L$ of $\tr^i_L$ and $\ac^i_R$ of $\tr^i_R$ at denoising step $i$, and (2) the left- and right-hand proprioceptive information $\hist_L$ and $\hist_R$, into 3D tokens.  We use the same 3D Relative Denoising Transformer architecture to contextualize all these tokens and predict the translation and rotation noise and the end-effector opening for both arms.

\paragraph{Flow Matching Action Prediction Objective:}
We replace the DDPM-based diffusion method with \textit{rectified flow}. The noisy left- and right-hand trajectory estimate $\tr^i_L=(1-i) \epsilon_L + i \tr_{t,L}^0$ and $\tr^i_R=(1-i) \epsilon_R + i \tr_{t,R}^0$ become the linear interpolation at denoising step $i$, where $\tr_{t,L}^0$ and $\tr_{t,R}^0$ denote the clean trajectory, and $\epsilon_L$ and $\epsilon_R$ denote the sampled noise for the left- and right-hand end effector.  In particular, our model takes as input two-hand trajectory estimate tokens $\tr^i=\{\tr_{L}^i,\tr_{R}^i\}$, proprioception tokens $\mathbf{\hist}=\{\hist_{L},\hist_{R}\}$, language tokens $l$, and scene tokens $\oo$; it outputs the left- and right-hand velocity field $v_{\theta, L}, v_{\theta, R}$ and gripper openness $f^{\mathrm{open}}_{\theta, L}, f^{\mathrm{open}}_{\theta, L}$, respectively.  We ignore time step $t$ to simplify notations.

During training, we sample a time step $t$ uniformly, denoising step $\id \sim \sigma(\gN(0, I))$ from a logit-normal distribution and noise $\epsilon_L \sim \gN(0, I), \epsilon_R \sim \gN(0, I)$ from Gaussian distribution.  We use the $L2$ loss to supervise the velocity field and binary cross-entropy (BCE) to supervise the end-effector opening.  By ignoring the notation of time step $t$, the  objective reads:
\begin{align}
    \mathcal{L}_\theta &=  \|\epsilon_L - \tr_{L}^0 -v_{\theta, L}(\oo, l, \mathbf{\hist},\tr^i,i)\|^2 + \|\epsilon_R - \tr_{R}^0 -v_{\theta, R}(\oo, l, \mathbf{\hist},\tr^i,i)\|^2 \\
    &+ \mathrm{BCE}(f_{\theta, L}^{\mathrm{\small open}}(\oo, l, \mathbf{\hist},\tr^i,i),\mathbf{a}_{1:T, L}^{\mathrm{\small open}}) + \mathrm{BCE}(f_{\theta, R}^{\mathrm{\small open}}(\oo, l, \mathbf{\hist},\tr^i,i),\mathbf{a}_{1:T, R}^{\mathrm{\small open}}) ,
    \nonumber
\end{align}

\paragraph{Accelerating Dataloading:}
We replace 3DDA's data loaders and optimize the keypose sampling during batching, data type conversion, and unprojection and augmentation operations. Specifically, we change the episodic loading to random keypose sampling. In more detail, the 3DDA codebase loads entire episodes, chunks them and concatenates chunks from different episodes to form a batch. We, on the other hand, sample keyposes across random episodes. To efficiently do this, we used the Python library zarr, to lazily load and access data indices across all episodes simultaneously. This offers the following advantages: i) we avoid loading whole episodes at once to only use a chunk, as this wastes time for data that is not used; ii) we ensure higher diversity in every batch; iii) we ensure a fixed batch size, contrary to 3DDA that concatenates chunks of possibly different sizes.

We handle data types to ensure faster batch collation. Specifically, we make sure to always load uint8 images and half-precision depth maps. This significantly speeds up batch formation, especially when large batch sizes are used. The data is converted to the desired type (float32) after being loaded to the GPU, where data-type conversions are much faster. In contrast, 3DDA loads and batches float32 tensors. We found that handling data types cuts down the loading time by half.

Lastly, we move depth unprojection to point cloud and augmentations to GPU. This offers two advantages: i) loading single-channel depth is much faster than three-channel point clouds; ii) it allows for faster, batched operations that are optimized on GPUs.

\paragraph{Faster point sampling} We adopt density-biased sampling (DBS)~\cite{palmer_dbs} to replace farthest point sampling (FPS)~\cite{eldar_fps}. In more detail, 3DDA employs FPS in the feature space to sparsify the scene tokens. FPS maintains a set of candidate points and a set of sampled points. Then, it iteratively samples the candidate point with maximum average distance from all sampled points. We replace this with DBS, which first estimates the sparsity of a neighborhood around a point as the average distance of the $k=8$ nearest neighbors of that point. Then, it promotes sampling in the sparser neighborhoods. A fast batched version of DBS can be implemented in pure PyTorch~\cite{paszke2019pytorchimperativestylehighperformance}.

\paragraph{Mixed-precision training:} We allow for autocasting operations to half precision when possible. This reduces the memory footprint of our model and allows for larger batch sizes.

\paragraph{Efficient attention implementation:} We use a modern PyTorch implementation of attention that runs optimized C++ code under the hood and is faster by an order of magnitude, especially when combined with large batch sizes and half precision.

\paragraph{CUDA graph compilation:} This technique is offered in modern PyTorch versions. The model is compiled as a graph of non-dynamic operations, allowing for optimizing all operations. Making \method{} compilable required refactoring changes over 3DDA, such as removal of logic branches, CPU operations, in-graph language tokenization and custom kernel operations such as FPS.

%%%%%%%%%%%%%%%%%%%%%%%%%%%%%%%%%%%%%%%
\section{Experiments}
\label{sec:experiment}

We evaluate \method{} on learning manipulation behaviors from demonstration in simulation and the real world. Specifically, we test \method{} on the PerAct2 bimanual manipulation benchmark (Section~\ref{sec:peract2}), on the PerAct (Section~\ref{subsec:exp_singlearm1}) and HiveFormer (Section~\ref{subsec:exp_singlearm2}) unimanual manipulation benchmarks, and on a suite of 10 real-world bimanual tasks using the Aloha robot platform (Section~\ref{subsec:exp_real}).

\subsection{Evaluation on the PerAct2 Bimanual Manipulation Benchmark}
\label{sec:peract2}
\label{subsec:exp_peract2}

\begin{table*}[t]
    % \vspace{-20pt}
    \centering
    \begin{adjustbox}{width=\textwidth}
    \tb{@{}l|c|c|c|c|c|c|c|c@{}}{1.0}{
    & multi-task & \cellcolor{ImportantColor}Avg. & push & lift & push & pick up & put item & put bottle \\
    & training & \cellcolor{ImportantColor}Success & box & ball & buttons & plate & into drawer & into fridge \\
    \midrule
    ACT~\cite{zhao2023learningfinegrainedbimanualmanipulation} & \xmark & 5.9 \cellcolor{ImportantColor} & 0 & 36 & 4 & 0 & 13 & 0 \\
    RVT-LF~\cite{goyal2023rvt,grotz2024peract2} & \xmark & 10.5\cellcolor{ImportantColor} & 52 & 17 & 39 & 3 & 10 & 0 \\
    PerAct-LF~\cite{shridhar2023perceiver,grotz2024peract2} & \xmark & 17.5\cellcolor{ImportantColor} & 57 & 40 & 10 & 2 & 27 & 0 \\
    PerAct$^2$~\cite{grotz2024peract2} & \xmark & 16.8\cellcolor{ImportantColor} & 6 & 50 & 47 & 4 & 10 & 3 \\
    DP3~\cite{Ze2024DP3} & \xmark & -\cellcolor{ImportantColor} & 56 & 64 & - & - & - & - \\
    KStarDiffuser~\cite{lv2025spatial} & \xmark & -\cellcolor{ImportantColor} & 83 & 98.7 & - & - & - & - \\
    PPI~\cite{yang2025gripperkeyposeobjectpointflow} & \xmark & -\cellcolor{ImportantColor} & \textbf{96.7} & 89.3 & - & - & 79.7 & - \\
    AnyBimanual~\cite{lu2024anybimanual} & \cmark & 32\cellcolor{ImportantColor} & 46 & 36 & 73 & 8 & - & 26 \\
    $\pi_0$-keypose~\cite{black2024pi_0} & \cmark & 43.7\cellcolor{ImportantColor} & 93 & 97 & 38 & 41 & 40 & 22 \\
    3DFA (ours) & \cmark & \textbf{85.1}\cellcolor{ImportantColor} & 92.7$_{\pm 0.47}$ & \textbf{99.7$_{\pm 0.47}$} & \textbf{92.7$_{\pm 1.89}$} & \textbf{69.7$_{\pm 12.6}$} & \textbf{93.0$_{\pm 2.83}$} & \textbf{89.3$_{\pm 1.89}$} \\
    \midrule\midrule
    & multi-task & handover & pick up & straighten & sweep & lift & handover& take tray\\
    & training & item & laptop & rope & dust & tray & item (easy) & out of oven \\
    \midrule
    ACT~\cite{zhao2023learningfinegrainedbimanualmanipulation} & \xmark  & 0 & 0 & 16 & 0 & 6 & 0 & 2\\
    RVT-LF~\cite{goyal2023rvt,grotz2024peract2} & \xmark & 0 & 3 & 3 & 0 & 6 & 0 & 3 \\
    PerAct-LF~\cite{shridhar2023perceiver,grotz2024peract2} & \xmark & 0 & 11 & 21 & 28 & 14 & 9 & 8 \\
    PerAct$^2$~\cite{grotz2024peract2} & \xmark & 11 & 12 & 24 & 0 & 1 & 41 & 9 \\
    DP3~\cite{Ze2024DP3} & \xmark & - & 6.3 & - & 1.7 & - & 0 & - \\
    KStarDiffuser~\cite{lv2025spatial} & \xmark & - & 43.7 & - & 89 & - & 27 & - \\
    PPI~\cite{yang2025gripperkeyposeobjectpointflow} & \xmark & - & 46.3 & - & 98.7 & 92 & 62.7 & - \\
    AnyBimanual~\cite{lu2024anybimanual} & \cmark & 15 & 7 & 24 & 67 & 14 & 44 & 24 \\
    $\pi_0$-keypose~\cite{black2024pi_0} & \cmark & 2 & 27 & 7 & 2 & 72 & 59 & 68 \\
    3DFA (ours) & \cmark & \textbf{89.0$_{\pm 7.12}$} & \textbf{74.0$_{\pm 8.96}$} & \textbf{40.7$_{\pm 1.89}$} & \textbf{99.3}$_{\pm 0.47}$ & \textbf{94.7$_{\pm 0.47}$} & \textbf{96.0$_{\pm 5.65}$} & \textbf{94.7$_{\pm 1.89}$} \\
    \bottomrule    }
    \end{adjustbox}
    \caption{\textbf{Evaluation on PerAct2 bimanual benchmark}.  
    \method{}, AnyBimanual and  $\pi_0$-keypose are multi-task policies  and evaluate one checkpoint across all tasks, while others are single-task policies  and report results from the best checkpoint per task. \textbf{\method{} outperforms all prior arts by a large margin.}
    }
    \label{tab:peract2}
    \vspace{-10pt}
\end{table*}

PerAct2~\cite{grotz2024peract2} is a simulation benchmark for bimanual manipulation using a dual-arm setup with two Franka Emika Panda robots. The benchmark includes a suite of 13 bimanual tasks, each with 1 to 5 variations involving changes in object pose, appearance, or semantics (see Section~\ref{sec:rlbench_tasks} for details). For each task, the dataset provides 100 demonstrations for training and 100 episodes for evaluation. Methods are trained to predict the next end-effector keypose~\cite{grotz2024peract2}, which is then passed to an RRT-based motion planner~\cite{kuffner2000rrt} to generate a feasible joint trajectory from the current configuration to the predicted pose. 
PerAct2 supports five calibrated RGB-D cameras—front, left wrist, right wrist, left shoulder, and right shoulder—all capturing images at a resolution of $256 \times 256$. We find that using only the front, left wrist and right wrist is sufficient for all tasks. 
Task success rate is used as the primary evaluation metric.

\paragraph{Baselines}

We compare our method against several strong baselines, including: ACT~\cite{zhao2023learningfinegrainedbimanualmanipulation}, RVT-LF~\cite{goyal2023rvt,grotz2024peract2}, PerAct-LF~\cite{shridhar2023perceiver,grotz2024peract2}, and PerAct$^2$\cite{grotz2024peract2} (results taken from \cite{grotz2024peract2}); AnyBimanual\cite{lu2024anybimanual}; 3D Diffusion Policy (DP3)\cite{Ze2024DP3} and KStarDiffuser\cite{lv2025spatial} (results reported in \cite{lv2025spatial}); PPI~\cite{yang2025gripperkeyposeobjectpointflow}; and $\pi_0$\cite{black2024pi_0}. Notably, $\pi_0$ is a 2D robot policy pretrained on 10,000 hours of robot demonstration data and capable of both unimanual and bimanual manipulation. It takes visual input from three camera views (front and wrists) and outputs future joint angle trajectories. We adapt $\pi_0$ to predict 3D end-effector keyposes, which is the standard output format used across all policies in this setting and significantly improves its performance. We call this variant $\pi_0$-keypose. Our method, along with $\pi_0$-keypose and AnyBimanual, uses \textit{ multitask training} and evaluates performance using the final checkpoint, whereas other baselines train separate models per task and report results from the best intermediate checkpoint for each task. Second, several baselines—including AnyBimanual, DP3, PPI, and KStarDiffuser—are evaluated on \textit{only a subset} of the 13 benchmark tasks. For further details on all baselines, please refer to Section~\ref{sec:detailed_baselines}.

\paragraph{Results} We show quantitative results in Table~\ref{tab:peract2}. \method{} largely outperforms all baselines, with an absolute improvement of 68.3\% over PerAct$^2$. When isolating the same 5 tasks in which both PPI and KStarDiffuser report results, \method{} achieves 92.3\%, outperforming them by 13.6\% and 24\% absolute respectively. Notably, our method with 3.8M parameters outperforms $\pi_0$ which has nearly 1000 times more parameters. We show that, although large-scale pretraining is useful, explicitly incorporating 3D information into the model is a strong inductive bias.  We also test 3DDA on this benchmark, which also achieves an average success rate of 85\%. We discuss failure cases in Section~\ref{sec:failure}.

% \begin{table}[!h]
% \begin{minipage}[t]{0.3\linewidth}
%     \includegraphics[height=\linewidth]{figs/sampling_curve}
% \end{minipage}
% \hfill
% \begin{minipage}[t]{0.3\linewidth}
%     \includegraphics[height=\linewidth]{figs/point_sampling}
% \end{minipage}
% \hfill
% \begin{minipage}[t]{0.3\linewidth}
%     \includegraphics[height=\linewidth]{figs/training_time}
% \end{minipage}
% \end{table}

\begin{figure*}[t]
    % \vspace{-25pt}
    \imwjpg{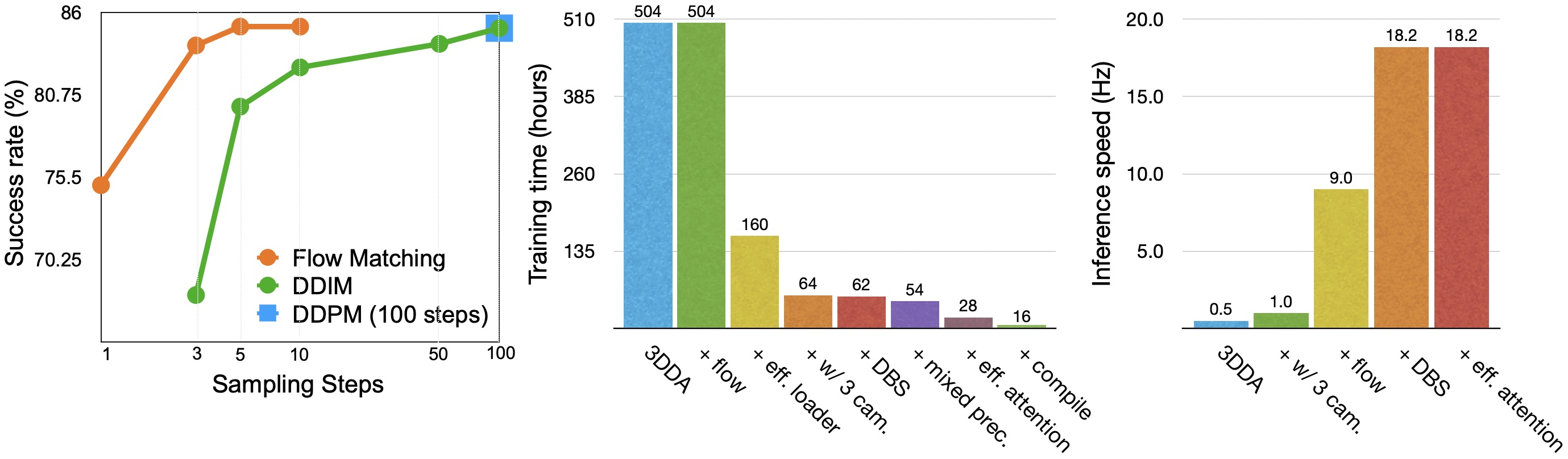}{1.0}
    \caption{\textbf{Ablation study} on PerAct2. \textbf{Left}: \method{}'s performance is stable even with as few as three denoising steps, contrary to the variants that use DDIM and DDPM. \textbf{Middle}: all design choices significantly contribute to lowering the training time from 504h to 16h (4 L40S GPUs). \textbf{Right}: Our contributions drastically improve inference speed from 0.5Hz to 18.2Hz (on one L40S GPU).}
    \label{fig:ablation}
    \vspace{-10pt}
\end{figure*}

\paragraph{Ablations}
We compare different denoising methods and quantify the effect of each contribution on training time and control frequency (Figure \ref{fig:ablation}). Key observations:

\begin{itemize}
    \item \textbf{Denoising method impact}: DDPM is the least flexible, requiring 100 denoising steps to achieve 85.0\% SR. DDIM~\cite{DDIM} reduces steps but suffers a sharp performance drop. Flow matching is the most robust, maintaining full 85.1\% performance even with 5 steps, 83.9\% with 3 steps, and 75.2\% with just 1 step.

    \item \textbf{Training time reduction}: Starting from the original 3DDA (adapted for bimanual manipulation), replacing its data-loading scheme with ours cuts training time by two-thirds. Additional optimizations cumulatively yield a 10× speedup, for a \textbf{total 30× training speedup}.

    \item \textbf{Control frequency gains}: Switching from DDPM to flow matching reduces denoising steps by 20× but raises control frequency only 9×, as scene encoding dominates the forward pass. Replacing FPS with DBS halves this cost, doubling control frequency. The attention implementation has negligible effect at test time, as it is tuned for large batch sizes.
\end{itemize}

%Third, we test the impact of wrist cameras in Figure~\ref{fig:ablation}c. We implement a single-camera version of \method{} that uses only the front camera, with all other hyperparameters fixed. This model achieves 81.1\% and runs at 66.7Hz. Although in many tasks the performance is similar to the 3-camera \method{}, there are tasks such as ``push buttons'' where the gap is significant, due to the importance of first-person observations for seeing and reaching the relevant objects.

%Lastly, in Figure~\ref{fig:ablation}d, we compare the training time of \method{} to that of the original implementation of 3DDA when trivially adapted for bimanual manipulation. Both trained on 4 A100 GPUs, 3DDA takes 21 days, \method{} only needs 30 hours. Notably, the training time can be significantly reduced when combined with CUDA graph compilation (\texttt{torch.compile}) or using fewer cameras, options not applicable to the original 3DDA implementation. 

\subsection{Evaluation on the PerAct  Unimanual Manipulation Benchmark}
\label{subsec:exp_singlearm1}

We evaluate our method on the 18-task PerAct benchmark~\cite{shridhar2023perceiver} to enable a direct comparison with 3DDA~\cite{ke20243d} in terms of both performance and efficiency. All policies are trained in a multi-task setting using 100 demonstrations per task and evaluated over 25 episodes per task. The setup includes four calibrated RGB-D cameras: front, wrist, left shoulder, and right shoulder (task details in Section~\ref{sec:peract_tasks}).

We compare two variants of \method{} against 3DDA in Figure~\ref{fig:peract_hiveformer}, with all models trained to predict keyposes. When using all four cameras, \method{} achieves performance on par with 3DDA while offering a 28× faster inference speed and requiring 6× less training time. With only two cameras (front and wrist), \method{} further reduces computational cost—14× less training time and 30× faster inference—while incurring only a minor drop in performance. Full results are provided in Section~\ref{sec:detailed_peract_results}.

\subsection{Evaluation on the  HiveFormer Unimanual Manipulation Benchmark}
\label{subsec:exp_singlearm2}

We evaluate \method{} on the 74-task HiveFormer benchmark~\cite{hiveformer} to showcase its ability to predict dense end-effector trajectories rather than sparse keyposes. This capability enables planner-free execution, which is critical for tasks requiring continuous interaction with objects and the environment. For instance, when opening a fridge, the end-effector must follow the door’s rotational arc to respect its mechanical limits—a constraint that an RRT planner does not account for.

All policies are trained on one task at a time with 100 demonstrations and evaluated on 100 test episodes. We report results across the full 74-task benchmark and also highlight performance on a subset of 8 challenging tasks~\cite{xian2023chaineddiffuser,chisari2024learning} that demand continuous interaction and cannot be solved by simply predicting keyposes. Further details are provided in Sections~\ref{sec:hiveformer_tasks} and~\ref{sec:chaineddiffuser_tasks}.

 %We employ the 74-task benchmark of HiveFormer~\cite{hiveformer} to demonstrate the ability of \method{} to predict dense trajectories instead of sparse keyposes. This allows for planner-free execution, which is essential for tasks that require continuous interaction with the objects and the environment. For example, when opening a fridge, the end-effector trajectory should follow the door’s rotation arc to respect its mechanical limits, a constraint that the RRT planner does not consider. Policies are trained on one task at a time, using 100 demos, and tested on 100 episodes. We compare our performance on all 74 tasks and also separately consider a subset of 8 challenging tasks \cite{xian2023chaineddiffuser,chisari2024learning} that require continuous interaction with the environment and thus cannot be solved by simply predicting keyposes- more in Sections~\ref{sec:hiveformer_tasks},~\ref{sec:chaineddiffuser_tasks}.

\paragraph{Baselines} We compare against: (1) keypose-prediction models HiveFormer~\cite{hiveformer}, InstructRL~\cite{liu2022instruction} and Act3D~\cite{gervet2023act3d}, which use three cameras; (2) close-loop trajectory-prediction models PointFlowMatch~\cite{chisari2024learning} and DP3~\cite{Ze2024DP3}, which use five cameras; (3) ChainedDiffuser~\cite{xian2023chaineddiffuser}, a two-stage policy consisting of a keypose predictor and a keypose-conditioned trajectory predictor, using three cameras.

\method{} is trained to jointly predict the next end-effector keypose and the dense trajectory until the next keypose in a non-hierarchical, single-forward-pass fashion. It relies on two camera observations, front and wrist, to maintain consistency with the rest of our experiments. More details in Section~\ref{sec:detailed_baselines}.

\begin{figure*}[t]
    % \vspace{-25pt}
    \imwjpg{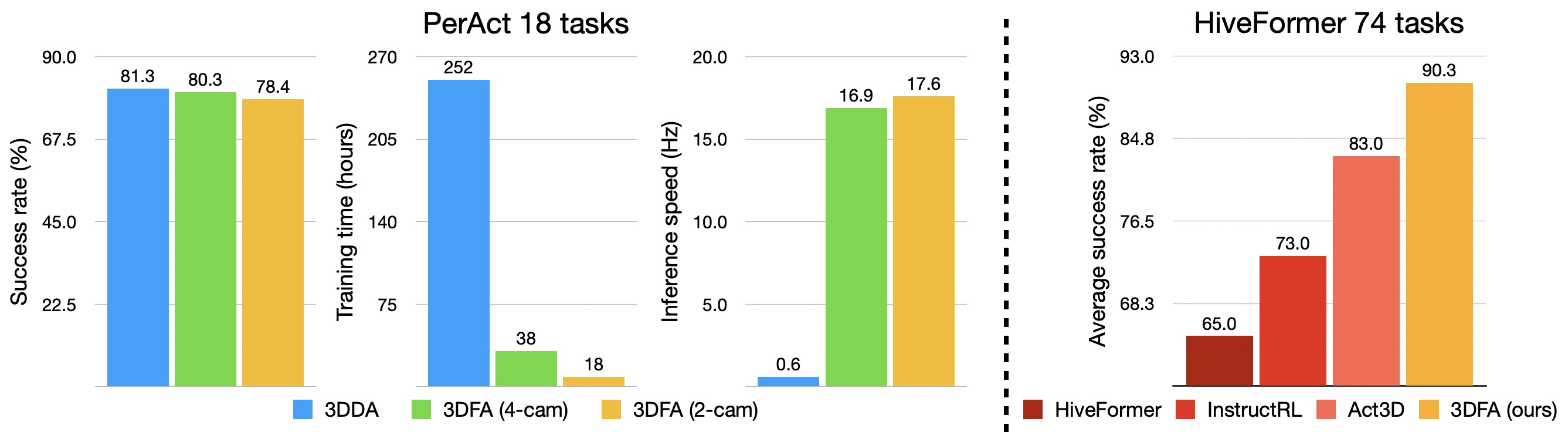}{1.0}
    \caption{\textbf{Single-arm manipulation results.} \textbf{Left}: On the PerAct 18-task benchmark, \method{} matches 3DDA's performance while reducing training time by over 6× and achieving 28× faster inference. \textbf{Right}: \method{} achieves a new state-of-the-art on the 74 tasks of HiveFormer.}
    \label{fig:peract_hiveformer}
    % \vspace÷{-5pt}
\end{figure*}
\begin{table*}[t]
    \centering
    \begin{adjustbox}{width=0.85\textwidth}
    \tb{@{}l|c|c|c|c|c|c|c|c|c|c@{}}{0.5}{
    & \cellcolor{ImportantColor}Avg. & Num. & unplug & close & open & open & frame off & open & books in & shoes in \\
    & \cellcolor{ImportantColor}Success & stages & charger & door & box & fridge & hanger & oven & shelf & box \\
    \midrule
    DP3~\cite{Ze2024DP3} & 28.5 \cellcolor{ImportantColor} & 1 & 33.3 & 76.0 & 98.3 & 4.3 & 12.3 & 0.3 & 3.7 & 0.0 \\
    PointFlowMatch~\cite{chisari2024learning} & 67.8\cellcolor{ImportantColor} & 1 & 83.6 & 68.3 & 99.4 & 31.9 & 38.6 & 75.9 & 68.8 & 76.0 \\
    ChainedDiffuser~\cite{xian2023chaineddiffuser} & 84.5\cellcolor{ImportantColor} & 2 & 95.0 & 76.0 & 96.0 & 68.0 & 85.0 & 86.0 & 92.0 & \textbf{78.0} \\

    \method{} (ours) & \textbf{91.3}\cellcolor{ImportantColor} & 1 & \textbf{99.0} & \textbf{96.0} & \textbf{100.0} & \textbf{70.0} & \textbf{96.0} & \textbf{99.0}& \textbf{99.0} & 71.0 \\
    }
    \end{adjustbox}
    \caption{\textbf{Evaluation on 8 RLBench tasks} from \cite{xian2023chaineddiffuser} that require continuous interaction with the environment.
    \method{} predicts dense trajectories and outperforms all baselines by a large margin.}
    \label{tab:pointflowmatch}
\end{table*}

We present results across all 74 tasks in Figure~\ref{fig:peract_hiveformer}. \method{} sets a new state of the art with a success rate of 90.3\%, surpassing Act3D by 7.3\%. On eight particularly challenging tasks (Table~\ref{tab:pointflowmatch}), our method outperforms the two-stage ChainedDiffuser by 6.8\% and PointFlowMatch by 23.5\% absolute, demonstrating its strong capability for continuous trajectory prediction. A detailed performance breakdown for all tasks is provided in Section~\ref{sec:detailed_hiveformer_results}.

\subsection{Evaluation in the Real World}
\label{subsec:exp_real}

%%%%%%%%%%%%%%%%%%%%%%%%%%%%%%% head view
% \begin{figure*}[h]
%     \resizebox{\textwidth}{!}{%
%     \small
%     \tb{@{}cccccc@{}}{0.1}{
%         \imw{figs/realworld_tasks/lift_ball}{0.2} & \imw{figs/realworld_tasks/straighten_rope}{0.2} & \imw{figs/realworld_tasks/pickup_plate}{0.2} & \imw{figs/realworld_tasks/stack_bowls}{0.2} & \imw{figs/realworld_tasks/insert_marker_into_cup}{0.2}\\
%         (a) Lift ball & (b) Straighten rope & (c) Pick up plate & (d) Stack bowls & (e) Put marker in cup \\
%         \imw{figs/realworld_tasks/handover_block}{0.2} & \imw{figs/realworld_tasks/stack_blocks}{0.2} & \imw{figs/realworld_tasks/open_marker}{0.2} & \imw{figs/realworld_tasks/close_ziploc}{0.2} & \imw{figs/realworld_tasks/insert_battery}{0.2}\\
%         (f) Hand over block & (g) Stack blocks & (h) Open marker & (i) Close ziploc & (j) Insert battery \\
%     }
%     }
%     \caption{\textbf{Real-world benchmark.} Our real-world benchmark has a suite of 10 bimanual tasks, including 5 easy tasks (top row) and 5 difficult tasks (bottom row).  Difficult tasks require highly precise and synergetic two-hand actions.  Mobile Aloha~\cite{fu2024mobilealohalearningbimanual} is adopted as the two-hand robot system.}
%     \label{fig:realworld_task}
% \end{figure*}

\begin{figure*}[t]
    % \vspace{-15pt}
    \resizebox{\textwidth}{!}{%
    \small
    \tb{@{}cccccc@{}}{0.1}{
        \imw{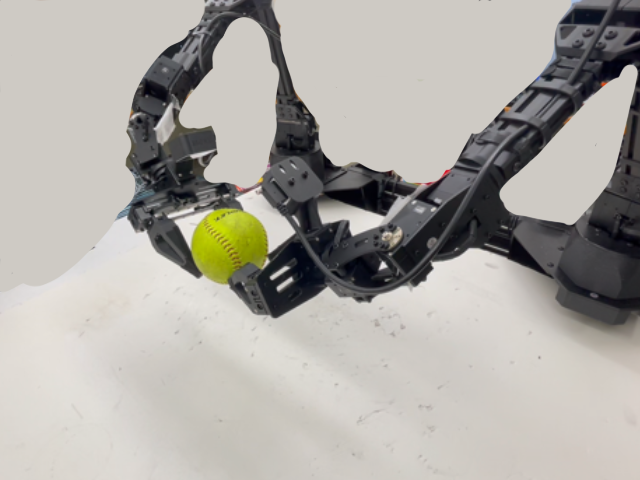}{0.2} & \imw{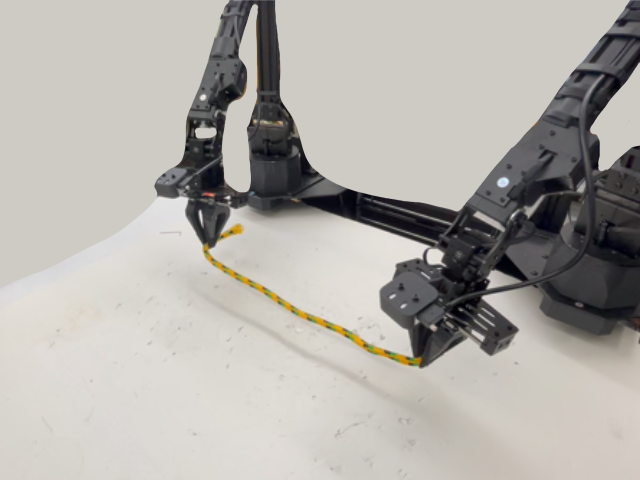}{0.2} & \imw{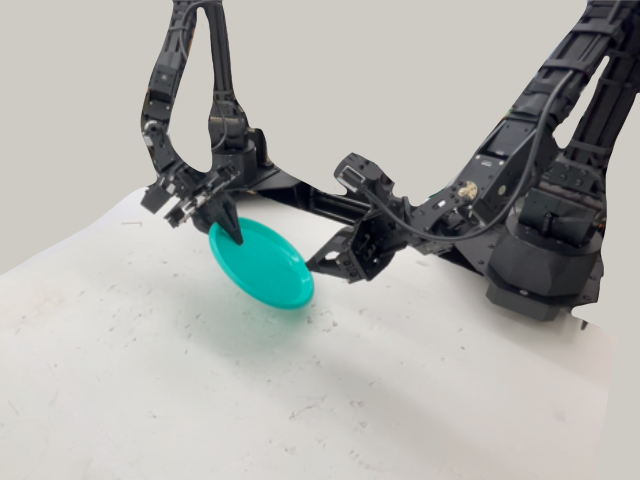}{0.2} & \imw{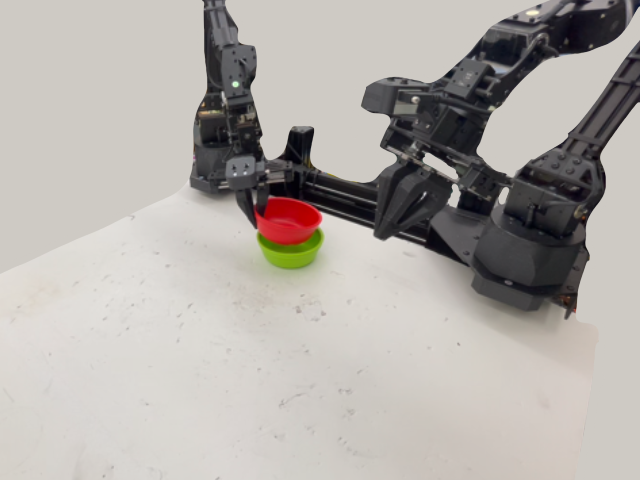}{0.2} & \imw{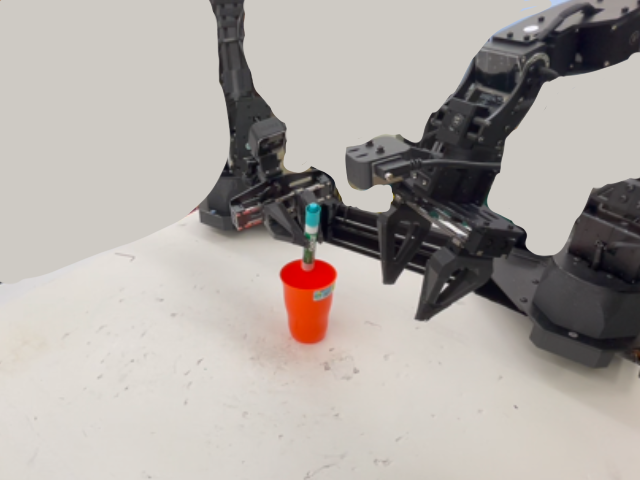}{0.2}\\
        (a) Lift ball & (b) Straighten rope & (c) Pick up plate & (d) Stack bowls & (e) Put marker in cup \\
        \imw{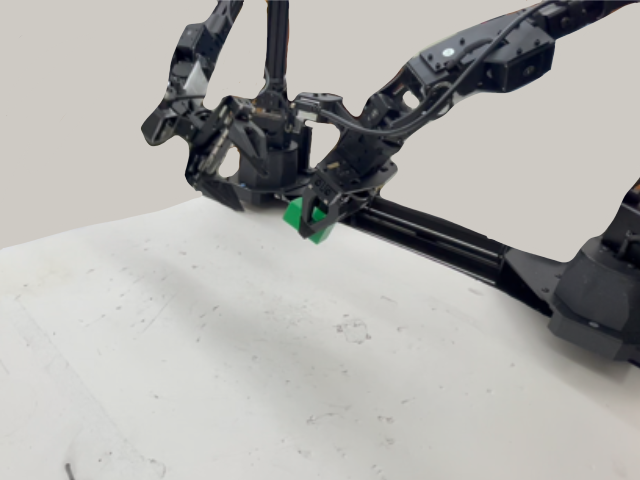}{0.2} & \imw{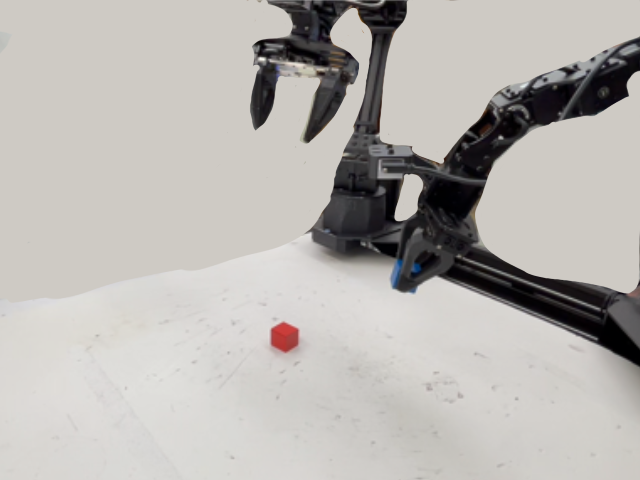}{0.2} & \imw{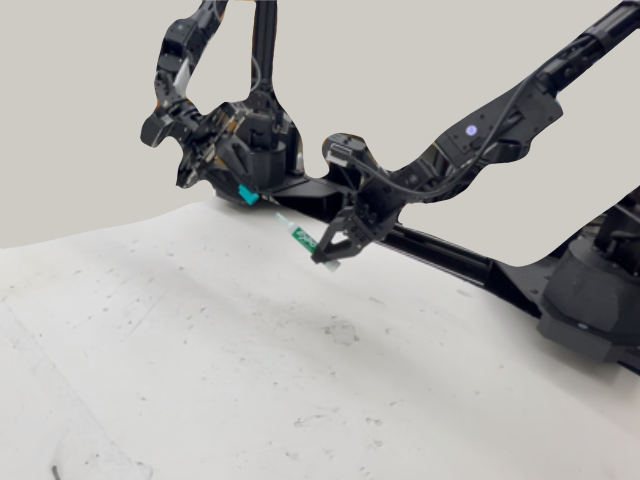}{0.2} & \imw{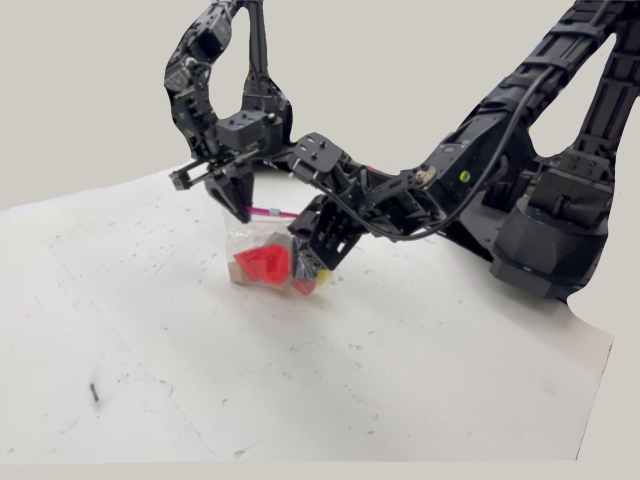}{0.2} & \imw{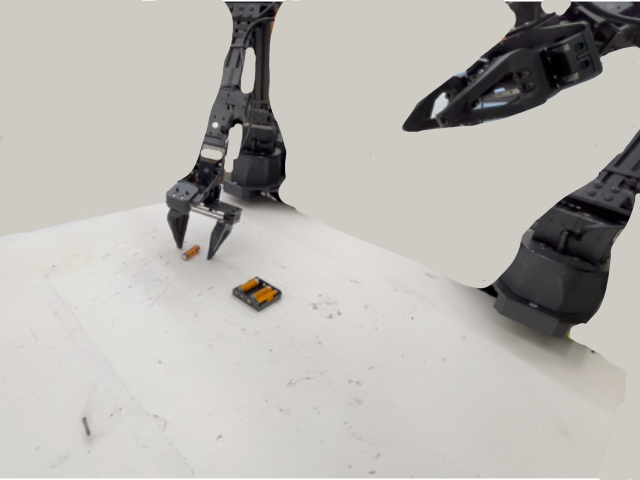}{0.2}\\
        (f) Hand over block & (g) Stack blocks & (h) Open marker & (i) Close ziploc & (j) Insert battery \\
    }
    }
    \caption{\textbf{Real-world benchmark.} Our real-world benchmark consists of 10 bimanual tasks, divided into 5 easy tasks (top row) and 5 difficult tasks (bottom row).  %Difficult tasks require highly precise and synergistic two-hand actions.  Mobile Aloha~\cite{fu2024mobilealohalearningbimanual} is adopted as the two-arm robot system.}
    }
    \label{fig:realworld_task}
    % \vspace{-15pt}
\end{figure*}

We construct a challenging real-world multi-task manipulation benchmark using Mobile Aloha~\cite{fu2024mobile}, a dual-arm mobile manipulator equipped with a front-facing ZEDX RGB-D camera and two wrist-mounted RealSense D405 RGB-D sensors. Our benchmark (Figure~\ref{fig:realworld_task}, Section~\ref{sec:real_world_tasks}) comprises 10 tasks that  demand precise and coordinated two-handed manipulation, examples include \textit{open marker}, \textit{close ziploc}, and \textit{insert battery}, surpassing the complexity of PerAct2 tasks.  We collect 40 demonstrations per task,  recording visual observations and joint actions at 5 Hz. All models are trained or fine-tuned using the same demonstration set and evaluated on 20 episodes per task.

\paragraph{Baselines}
We compare against two strong baselines:  (1) $\pi_0$~\cite{black2024pi_0}, a generalist 2D manipulation policy, and (2) iDP3~\cite{ze2024generalizable}, a variant of DP3 adapted for dual-arm humanoid systems with architectural enhancements.  All models, including ours, are trained for closed-loop trajectory prediction without intermediate keypose supervision.  Following their original designs, both baselines directly predict joint angles for the robot arms, whereas \method{} predicts 3D end-effector poses, which are then converted to joint commands via inverse kinematics. To mimic human operation, all models output joint values for the Aloha leader arms, with the follower arms mirroring the motion. \method{} and iDP3 are 3D policies that require accurate depth sensing; due to high noise from the wrist-mounted depth sensors, we exclude wrist camera inputs for both. In contrast, the 2D $\pi_0$ policy is depth-independent and utilizes all three camera views. All image inputs are downsampled to a resolution of $256 \times 256$.

%Following the original papers, both baselines directly predict joint angles of the robot arms, whereas \method{} predicts 3D end-effector poses that are converted to joint commands via inverse kinematics. To replicate human operation, all models output joint values for the Aloha leader arms, with the follower arms mirroring the motion. \method{} and iDP3 are 3D policies and rely on accurate depth information. Due to the high noise in the wrist-mounted depth sensors, we exclude wrist camera inputs from both methods. In contrast, the 2D policy $\pi_0$ is not dependent on depth data and uses all three camera views. All image inputs are downsampled to $256 \times 256$ resolution.

\begin{table*}[t]
    \centering
    % \vspace{-5pt}
    \begin{adjustbox}{width=\textwidth}
    \tb{@{}l|c|c|c|c|c|c|c|c|c|c|c|c|c@{}}{0.2}{
    % & multi-task & \cellcolor{ImportantColor}Avg. & lift & straighten & pick up & stack & put marker & handover & stack & open & close & insert\\
    % & training & \cellcolor{ImportantColor}Success & ball & rope & plate & bowl & into bowl & block & blocks & marker & ziploc & battery\\
    % \midrule
    % Pi0 & \cmark & TBD \cellcolor{ImportantColor} & TBD & 40 & 10 & TBD & TBD & TBD & TBD & TBD & TBD & TBD \\
    % iDP3 & \xmark & 24.5\cellcolor{ImportantColor} & 45 & 35 & 30 & 40 & 35 & 25 & 15 & 10 & 10 & 0\\
    % Bi3DDA (ours) & \cmark & \textbf{58}\cellcolor{ImportantColor} & \textbf{90} & \textbf{85} & \textbf{75} & \textbf{85} & \textbf{80} & \textbf{60} & \textbf{55} & \textbf{20} & \textbf{20} & \textbf{10}\\
    % 3DFA (front) & \cmark & \textbf{53.5}\cellcolor{ImportantColor} & \textbf{85} & \textbf{75} & \textbf{80} & \textbf{80} & \textbf{70}& \textbf{60} & \textbf{45} & \textbf{25} & \textbf{10} & \textbf{5} \\
    & \cellcolor{ImportantColor}Avg. & Inf. & Params & lift & straighten & pick up & stack & put marker & handover & stack & open & close & insert\\
    & \cellcolor{ImportantColor}Success & speed & in M & ball & rope & plate & bowl & into bowl & block & blocks & marker & ziploc & battery\\
    \midrule
    $\pi_0$~\cite{black2024pi_0} & 32.5 \cellcolor{ImportantColor} &100ms & 3238 & 85 & 75 & 40 & 20 & 20 & 15 & 35 & 20 & 10 & 5 \\
    iDP3$^\dagger$~\cite{ze2024generalizable} & 24.5\cellcolor{ImportantColor} & 420ms & 68.8 & 45 & 35 & 30 & 40 & 35 & 25 & 15 & 10 & 10 & 0\\

    % Bi3DDA: 30 denoise steps 640 ms
    % Bi3DDA (ours)& \textbf{58}\cellcolor{ImportantColor} & & \textbf{90} & \textbf{85} & \textbf{75} & \textbf{85} & \textbf{80} & \textbf{60} & \textbf{55} & \textbf{20} & \textbf{20} & \textbf{10}\\
    \method{} (ours) & \textbf{53.5}\cellcolor{ImportantColor} & 54ms & 3.8 & \textbf{85} & \textbf{75} & \textbf{80} & \textbf{80} & \textbf{70}& \textbf{60} & \textbf{45} & \textbf{25} & \textbf{10} & \textbf{5} \\
    }
    \end{adjustbox}
    \caption{\textbf{Evaluation in the real world.} 
    \method{} and $\pi_0$ adopt multi-task learning settings, while iDP3 uses a single-task learning setup. \textbf{\method{} outperforms $\mathbf{\pi_0}$ and iDP3 on all tasks by a large margin}.
    }
    \label{tab:realworld}
    \vspace{-10pt}
\end{table*}

\paragraph{Results}
We show the quantitative results in Table~\ref{tab:realworld}.  \method{} outperforms $\pi_0$ and iDP3 on most tasks.  We find that $\pi_0$ is sensitive to occlusion and cannot locate and reach the object if it is not visible in the wrist observations. We also find that iDP3 struggles to predict precise end-effector poses, as the robot often approaches but fails to grasp the object.  In contrast, our method consistently outperforms these baselines in most tasks. We refer to Sections~\ref{sec:detailed_rw_results},~\ref{sec:failure} for more analysis.

\subsection{Limitations and Future Directions}

While \method{} significantly reduces training and inference costs, it retains a key limitation: as a 3D policy, it depends on accurate depth sensing and camera calibration—resources often unavailable in large-scale, real-world imitation learning datasets. In particular, wrist-mounted cameras present the greatest calibration challenges. We are exploring approaches to relax these depth and calibration requirements, including architectures capable of jointly processing 2D and 3D observations~\cite{jain2024odinsinglemodel2d}, as well as leveraging recent advances in metric depth estimation and automatic calibration from the broader computer vision community. 

The compact parameterization of \method{}, while contributing to its efficiency, also makes it susceptible to certain errors. For instance, on PerAct, the model may struggle with unseen variations encountered at test time. In our real-world benchmark, it has difficulty with tasks demanding fine force control or extreme precision. Scaling up the model’s capacity and incorporating strong vision-language models for richer visual and language understanding are promising directions for addressing these challenges.

%Despite tremendous improvements in training and inference cost, \method{} still has an important limitation: as a 3D policy, it requires accurate depth sensors and camera calibration, often not available in large-scale real-world imitation learning datasets. We are working on removing the depth and calibration assumptions from wrist cameras, which are inherently the most difficult to calibrate, and building architectures that can absorb both 2D and 3D observations~\cite{jain2024odinsinglemodel2d} or leverage computer vision breakthroughs for metric depth estimation and auto-calibration.

%Due to its low parameter count, \method{} can be prone to various sources of errors. On PerAct for example, the model struggles with understanding the (sometimes unseen) variations at test time. In our real-world benchmark, it struggles with tasks that require force control or very high precision. Scaling the parameters of \method{} and employing strong VLMs for visual and language understanding are clear avenues for future work. 

%Next, \method{} is still trained with imitation learning, thus it cannot easily discover a new behavior on its own. Given their data efficiency, introducing 3D policies into a reinforcement learning framework could dramatically speed up exploration and data collection.

%

\section{Conclusion}
\label{sec:summary}
We introduced \methodfull{} (\method{}), a fast and versatile 3D manipulation policy that combines flow matching with pretrained 3D scene representations. Through targeted architectural and system-level optimizations, \method{} achieves over 30X faster training and inference than prior 3D diffusion-based policies, while setting new state of the art on both bimanual (PerAct2) and unimanual (RLBench-74) benchmarks. It delivers real-time performance, scales to real-world tasks, and removes the need for motion planning via direct dense trajectory prediction. These results position \method{} as an efficient and general-purpose framework for unimanual and bimanual  robot manipulation.

%We present \methodfull{}, a 3D policy for unimanual and bimanual  robot manipulation that reasons jointly across visual and action tokens in a unified 3D space.  We present \methodfull{}, a policy that extends 3D Diffuser Actor to bimanual manipulation and dramatically accelerates its training and inference speed.  Our method sets a new state of the art on the bimanual manipulation benchmark PerAct2 and a real-world benchmark, significantly outperforming much larger foundational policies. Our innovations are also effective in single-arm manipulation, where \method{} exhibits vast speedups over 3DDA on PerAct, together with a new state of the art on 74 RLBench tasks, also demonstrating planner-free execution capabilities. Reducing training/inference times makes \method{} a drop-in replacement for 3DDA that is far more suitable for real-world applications. Our work is a clear demonstration that even in the era of large-scale VLAs, policies that leverage 3D scene representations are still relevant. As concurrent work in the vision community continues to increase the amount of well-calibrated robotics data, the improvements introduced in this paper will only become more important.
% An exciting future direction is to scale up our policy by training on large-scale real-world datasets and to further enhance spatial and dynamic reasoning for contact-rich tasks. 
% \input{7_misc}

\newpage
%%%%%%%%% REFERENCES
\bibliography{bibs/egbib,bibs/refs,bibs/act3drefs,bibs/diffusion,bibs/flow_matching,bibs/neuripsrefs,bibs/bimanual}

\clearpage
\hypersetup{linkcolor=black}
\etocdepthtag.toc{mtappendix}
\etocsettagdepth{mtchapter}{none}
\etocsettagdepth{mtappendix}{subsection}
\tableofcontents
\hypersetup{linkcolor=red}
\appendix
\clearpage

\section{Related Work}
\label{sec:work}
\noindent\textbf{Bimanual manipulation.} 
Bimanual manipulation is challenging due to the need for precise coordination between two arms. A key bottleneck is the difficulty of collecting large-scale, high-quality bimanual demonstration data, which has historically constrained the scalability of approaches~\cite{Grannen2022LearningBS,grannen2023stabilizeactlearningcoordinate}. Recent works~\cite{zhao2023learningfinegrainedbimanualmanipulation,fu2024mobilealohalearningbimanual} have introduced more cost-effective pipelines and platforms for scaling real-world data collection. However, these methods primarily rely on RGB inputs and struggle to generalize across diverse tasks, object types, scene configurations or robot embodiments. To address these limitations, several multi-task simulation benchmarks have been proposed to facilitate large-scale demonstration collection and policy evaluation. PerAct2~\cite{grotz2024peract2} extends the RLBench~\cite{james2020rlbench} benchmark to support multi-task bimanual manipulation, with expert demonstrations generated using sampling-based motion planners~\cite{kuffner2000rrt}. RoboTwin~\cite{mu2025robotwindualarmrobotbenchmark} introduces a generative digital twin framework, leveraging 3D generative foundation models and large language models to create diverse expert datasets and real-world-aligned evaluation environments.

In parallel, the development of bimanual manipulation policies generally falls into two main categories. One line of research extends single-arm policy architectures to bimanual. For example, VoxAct-B~\cite{Liu2024VoxActBVA} builds upon the 3D voxel-based scene representations of \cite{shridhar2023perceiver} to jointly predict object and end-effector poses. Similarly, RDT-1B~\cite{liu2024rdt} follows the Vision-Language-Action (VLA) paradigm~\cite{brohan2023rt2,black2024pi_0,qu2025spatialvla}, fine-tuning large language models to improve generalization to diverse task instructions.  The second line of research composes multiple single-arm policies into a unified bimanual policy~\cite{lu2024anybimanual,motoda2025learning,jiang2025rethinking}. For instance, DIF~\cite{jiang2025rethinking} observes that bimanual manipulation involves both independent and coordinated actions between the two arms, so it proposes to learn separate single-arm policies and a coordination module to combine them for coordinated tasks. Our work belongs to the first category and extends the action space of \cite{ke20243d} to predict pose trajectories for both arms simultaneously.

\noindent\textbf{Diffusion and Flow Matching models in robotics.}
Diffusion models have emerged as powerful tools in imitation learning~\cite{chi2023diffusion,reuss2023goal,gkanatsios2023energy} and offline RL~\cite{chen2023offline,yang2024diffusiones,hansen2023idql}.  DDPM~\cite{DDPM} has been the most widely adopted diffusion algorithm to iteratively add and remove Gaussian noise to and from the samples, following conditional probability paths.  More recently, Flow Matching~\cite{liu2022flow,lipman2022flow} has drawn attention in robot learning.  Policies that use flow matching learn to predict the velocity field directly pointing toward the target action~\cite{black2024pi_0,funk2024actionflow,braun2024riemannian,zhang2024affordance,ding2024fast,zhang2024flowpolicy,chisari2024learning,wang2025flowram,bjorck2025gr00t,reussflower}, resulting in better sampling efficiency and lower computational cost than DDPM-based policies. \method{} also incorporates Flow Matching, yielding a state-of-the-art bimanual manipulation policy with real-time inference capabilities.

\section{Implementation details} \label{sec:hyper_parameter}
\paragraph{Architecture:} We closely follow 3DDA~\cite{ke20243d} in terms of architecture design. We use identical weights to 3DDA, with only the appropriate changes so that we handle and predict bimanual actions of 16 dimensions rather than 8. Specifically, this affects all layers that either project a noisy action token to a high-dimensional latent vector or predict an action token from a latent vector.

For completeness, we briefly describe the architecture here: \method{} uses CLIP~\cite{radford2021learning} to encode each image separately into a feature map, then assigns 3D positions to each 2D feature token using depth and the known camera parameters. The feature maps are merged into a single feature cloud, which is subsampled using density-biased sampling (DBS) to keep 1/4th of the tokens. The distance metric for DBS is computed in the feature space, not the 3D space. We call the subsampled feature cloud ``visual tokens".

The noisy trajectory estimate is encoded with a linear layer. Trajectory and visual tokens are concatenated in the sequence dimension and fed to a sequence of 6 self-attention layers. All tokens in this attention use 3D rotary positional embeddings, as used in \cite{ke20243d,gervet2023act3d}. Additionally, these layers are modulated by a signal computed by projecting i) the current end-effector pose and ii) the denoising step, into high-dimensional features. This signal is used to compute adaptive normalization parameters.

Lastly, MLPs are used on the output tokens to predict the translation velocity field, rotation velocity field and end-effector opening. Our model is language-conditioned, encodes the instruction with CLIP and then lets the visual and trajectory tokens attend to the language features, as in 3DDA.

\paragraph{Normalization of the output space:} 3D relative attention requires that the positional embeddings of all action tokens and visual tokens be expressed in a common coordinate frame. 
The original 3DDA normalizes the 3D space using the target end-effector’s statistics, which does not generalize to bimanual setups where each arm has distinct spatial distributions. Naively aggregating statistics across both arms and normalizing based on that would significantly compress the spatial distribution and harm performance. Instead, we compute positional embeddings in the global unnormalized frame, while still predicting trajectories in the per-arm normalized space. This hybrid approach preserves 3DDA’s attention mechanism while accommodating the bimanual setup.

\paragraph{Flow hyperparameters:} When training a Flow Matching model, we sample a denoising $i$ from a continuous internal between 0 and 1, where $i=0$ denotes pure noise and $i=1$ indicates a perfectly clear signal. 
One hyperparameter is how we sample $i$ during training. We experimented with uniform sampling, beta sampling (as in $\pi$0) and logit-normal sampling (as in \cite{reussflower}). We found beta sampling to consistently underperform the other two options, but little difference between uniform and logit-normal. We adopt logit-normal and sample $i$ as follows: we first sample x from a Gaussian with 0 mean and 1.5 std, then $i = \sigma(x)$, where $\sigma$ is the sigmoid function.

\paragraph{Other hyperparameters:} We follow the training hyperparameters of 3DDA~\cite{ke20243d}, in terms of optimizer selection, learning rate, etc. We use a constant batch size of 256 keyposes (64 per GPU) for PerAct2 and PerAct, 16 trajectories (one GPU) for the HiveFormer tasks, randomly sampled from different tasks and episodes. We found that our model converges in 300000 training steps on PerAct/PerAct2, contrary to 3DDA that needs 600000 steps. We hypothesize that this is due to the optimized batching scheme we propose. For the HiveFormer, we train single-task models until convergence, which usually takes less than 100000 steps or approximately 4 hours on one GPU.

\paragraph{Keypose discovery}
We use keyposes only on RLBench, while for our real-world experiments we predict trajectories directly. For our single-arm RLBench experiments, we use the heuristics from \cite{james2022q}, while for PerAct2 we use the ones from \cite{grotz2024peract2}, that adapts those of \cite{james2022q} for the bimanual setup. Specifically we examine each arm separately and a pose is marked as a keypose if (1) the end-effector state changes (grasp or release) or (2) the velocity magnitude approaches zero (often at pre-grasp poses or a new phase of a task). Once we compute a set keyposes for each arm, we then take the union of the two sets as the bimanual keyposes.

\section{Additional Experimental Results and Details}

\subsection{Peract2 tasks}
\label{sec:rlbench_tasks}
We provide brief descriptions of the 13 PerAct2 tasks for completeness. All tasks vary the object pose, appearance and semantics. For more details, we refer to the PerAct2 paper~\cite{grotz2024peract2}.  
\begin{enumerate}
    \item Push box: The scene is equipped with a heavy box and a target area. The robot needs to push the box using both arms to move it to the designated area. This task cannot be solved with one robot due to the weight of the box.

    \item Lift ball: The scene is equipped with a large ball. The robot needs to grasp the ball using both arms and lift it to a height above 0.95 m. This task is impossible to solve with one robot due to the size of the object.

    \item Push buttons: The scene contains three buttons of different colors. The robot needs to push two of the three buttons simultaneously, as specified in the language instruction.

    \item Pick up plate: The robot needs to pick and securely lift a plate. The robot must coordinate both arms to lift the plate without tilting or dropping it.

    \item Put item in drawer: The scene contains a furniture with drawers and a small item. The robot needs to open a specific drawer and place an item inside. The correct drawer is specified in the language instruction.

    \item Put bottle in fridge: The robot needs to open a fridge and put a bottle inside.

    \item Handover an item: There are multiple items of different colors on the table. The robot needs to pick one with one arm and hand it securely to the other arm. The correct item is specified by the language instruction.

    \item Pick up laptop: The robot needs to pick up a laptop that is placed on top of a block. This requires first moving the laptop into a position where it can be grasped and then lifting it.

    \item Straighten rope: The robot needs to straighten a rope so that both ends are placed in distinct target areas.

    \item Sweep dustpan: The scene is equipped with a broom, a dust pan, supporting objects and dust. The robot needs to sweep the dust into the dust pan using the broom. The task is considered successfully completed when all the dust is inside the dust pan.

    \item Lift tray: The robot needs to lift a tray for more than 1.2m. The tray is originally placed on a holder. An item is on top of the tray and must be balanced while the tray is lifted. The primary challenge lies in coordinating the lifting motion with both arms to maintain the balance of the item on the tray.

    \item Handover item (easy): This is a variant of the handover item task (described above), but the same object is always picked.

    \item Take tray out of oven: The robot needs to remove a tray that is located inside an oven. This involves opening the oven door and then grasping the tray.

\end{enumerate}

\subsection{Real-world tasks}
\label{sec:real_world_tasks}
We explain the real-world tasks and their success conditions in more detail.
\begin{enumerate}
    \item Lift ball: The robot needs to use both arms to stabilize and lift a small ball without grasping it.

    \item Straighten rope: The robot grasps both ends of the rope and pulls until the rope becomes straight.

    \item Pick up plate: The robot manipulates a small plate. The left arm needs to tilt the plate, while the right arm grasps it and lifts it up.

    \item Stack bowls: The robot needs to stack two bowls on top of each other.

    \item Put marker in cup: The robot needs to use the left arm to reach and fetch the cup, then use the right arm to grasp the marker and put it into the cup.
    
    \item Handover block: The robot uses its left arm to pick up a block and give it to the right arm.
    
    \item Stack blocks: The robot needs to stack two cubes (3cm on each side) on top of each other, which is harder than stacking bowls, since the blocks are smaller.
    
    \item Open marker: The robot needs to grasp the marker with its left arm and then use the right arm to grasp the cap and remove it from the marker.
    
    \item Close ziploc: The robot first grasps one end of the ziploc bag and then grasps the gray slider on the ziploc to close the bag.

    \item Insert battery: The robot moves its left arm near the slot for holding and then the right arm grasps the battery and inserts the battery into the slot. 
\end{enumerate}

The tasks ``stack bowls", ``put marker in cup" and ``stack blocks" are bimanual due to the reachability of the objects: the objects are initially placed in locations where only one of the two arms can reach them.

\subsection{PerAct tasks}
\label{sec:peract_tasks}
PerAct offers a suite of 18 tasks and 249 variations, making it one of the most challenging RLBench-based benchmarks. The variations include instance references, e.g. ``open the top/middle/bottom drawer", color, e.g. ``push the blue/purple/etc button", counting, e.g. ``stack 2/3 cups", and others such as object category, size and shape. We enumerate the 18 tasks here and refer to the PerAct paper~\cite{shridhar2023perceiver} for more details:  
\begin{enumerate}
    \item Open a drawer

    \item Slide a block to a colored zone

    \item Sweep the dust into a dustpan

    \item Take the meat off the grill frame

    \item Turn on the water tap

    \item Put a block in the drawer

    \item Close a jar

    \item Drag a block with the stick

    \item Stack blocks

    \item Screw a light bulb

    \item Put the cash in a safe

    \item Place a wine bottle on the rack

    \item Put groceries in the cardboard

    \item Put a block in the shape sorter

    \item Push a button

    \item Insert onto a peg

    \item Stack cups

    \item Hang cups on the rack
\end{enumerate}

\subsection{74 HiveFormer tasks}
\label{sec:hiveformer_tasks}
HiveFormer offers a suite of 74 tasks, grouped into 9 categories. The tasks come without variations, following \cite{hiveformer,gervet2023act3d}. We enumerate the groups and tasks here and refer to \cite{hiveformer,gervet2023act3d} for more details:

\begin{itemize}
    \item Planning (can be decomposed into multiple sub-goals): basketball in hoop, put rubbish in bin, meat off grill, meat on grill, change channel, tv on, tower3, push buttons, stack wine.

    \item Tools (need interaction with a target object): slide block to target, reach and drag, take frame off hanger, water plants, hang frame on hanger, scoop with spatula, place hanger on rack, move hanger, sweep to dustpan, take plate off colored dish rack, screw nail.

    \item Long term: wipe desk, stack blocks, take shoes out of box, slide cabinet open and place cups.

    \item Rotation-invariant: reach target, push button, lamp on, lamp off, push buttons, pick and lift, take lid off saucepan.

    \item Motion planner (require continuous interaction with the object/environment): toilet seat down, close laptop lid, open box, open drawer, close drawer, close box, phone on base, toilet seat up, put books on bookshelf.

    \item Multimodal (multiple solutions are possible): pick up cup, turn tap, lift numbered block, beat the buzz, stack cups.

    \item Precision (high-precision requirements): take usb out of computer, play jenga, insert onto square peg, take umbrella out of umbrella stand, insert usb in computer, straighten rope, pick and lift small, put knife on chopping board, place shape in shape sorter, take toilet roll off stand, put umbrella in umbrella stand, setup checkers.

    \item Screw (require (un)screwing an object): turn oven on, change clock, open window, open wine bottle.

    \item Visual Occlusion (large objects occlude certain views): close microwave, close fridge, close grill, open grill, unplug charger, press switch, take money out safe, open microwave, put money in safe, open door, close door, open fridge, open oven, plug charger in power supply
\end{itemize}

\subsection{8 ChainedDiffuser tasks}
\label{sec:chaineddiffuser_tasks}
While most of the 74 HiveFormer tasks can be solved by simply predicting keyposes and employing the RRT motion planner to obtain a trajectory to execute, there are several tasks that require continuous interaction with the objects and the environment. For example, when opening a fridge, the end-effector trajectory should follow the door’s rotation arc to respect its mechanical constraints. ChainedDiffuser~\cite{xian2023chaineddiffuser} identified 10 such challenging tasks. PointFlowMatch~\cite{chisari2024learning} further investigates trajectory prediction to solve 8 of those 10 tasks. To directly compare \method{}'s ability to predict dense trajectories against those works, we evaluate on the same 8 tasks that both works consider, namely:
\begin{enumerate}
    \item Unplug charger

    \item Close door

    \item Open box

    \item Open fridge

    \item Take frame off hanger

    \item Open oven

    \item Put books on bookshelf

    \item Take shoes out of box
\end{enumerate}

All these tasks require specific motion that cannot be modeled by an RRT planner that connects keyposes with collision-free linear segments. For example, unplugging a charger requires a smooth motion until the charger is out of the socket. Assuming that the end-effector has grasped the charger, the RRT trajectory is not guaranteed to produce a trajectory that smoothly extracts the charger from the socket.

\subsection{Additional details on baselines}
\label{sec:detailed_baselines}
On PerAct2, we compare against: (1) ACT~\cite{zhao2023learningfinegrainedbimanualmanipulation}, a 2D transformer architecture that is trained as a conditional VAE to predict a sequence of actions; (2) RVT-LF~\cite{goyal2023rvt,grotz2024peract2}, that unprojects 2D views to form a point cloud, renders virtual views and feeds them to a transformer to predict the 3D actions for each arm in sequence; (3) PerAct-LF~\cite{shridhar2023perceiver,grotz2024peract2}, that voxelizes the 3D space and uses a Perceiver~\cite{jaegle2021perceiver} architecture to predict the 3D actions for each arm in sequence; (4) PerAct$^2$~\cite{grotz2024peract2}, which shares the same architecture as PerAct-LF but predicts the actions for the two arms jointly; (5) AnyBimanual~\cite{lu2024anybimanual}, which combines and adapts two pre-trained single-arm PerAct~\cite{shridhar2023perceiver} policies; (6) 3D Diffusion Policy (DP3)~\cite{Ze2024DP3}, which encodes 3D scenes with a point cloud encoder and uses a diffusion UNet~\cite{ronneberger2015u} to predict the 3D actions; (7) KStar Diffuser~\cite{lv2025spatial}, a diffusion graph convolutional network that regularizes end-effector pose prediction with predicting body joint angles; (8) PPI~\cite{yang2025gripperkeyposeobjectpointflow}, a 3D diffusion policy that regularizes action prediction by tracking points sampled from objects of interest; (9) $\pi_0$~\cite{black2024pi_0}, a state-of-the-art 2D robot generalist that is pre-trained on 10,000 hours of robot data and capable of performing bimanual manipulation. We adapted $\pi_0$ to predict end-effector keyposes and fine-tuned it using three cameras.

On the 74 HiveFormer tasks, we compare against: (1) HiveFormer~\cite{hiveformer}, a 3D policy that predicts actions as offsets from the input point cloud; (2) InstructRL~\cite{liu2022instruction}, a 2D policy that leverages pre-trained vision and language encoders and regresses 6-DoF actions; (3) Act3D~\cite{gervet2023act3d}, a 3D policy that predicts the next action location by iterating between coarse-to-fine sampling and featurization; (4) ChainedDiffuser~\cite{xian2023chaineddiffuser}, a two-stage policy that employs Act3D for keypose prediction and a diffusion-based trajectory optimization model to connect the current pose to the next keypose; (5) PointFlowMatch~\cite{chisari2024learning}, a flow-based 3D policy that encodes the input point cloud into a single vector using PointNet~\cite{qi2017pointnetdeeplearningpoint} and (6) DP3~\cite{Ze2024DP3}, discussed in the previous paragraph.

\method{} is trained to predict a keypose-horizon trajectory: it jointly predicts the next end-effector keypose and the dense trajectory until the next keypose in a non-hierarchical, single-forward-pass fashion. HiveFormer, InstructRL and Act3D are trained to predict end-effector keyposes and then use RRT~\cite{kuffner2000rrt} to plan a trajectory. ChainedDiffuser is a two-stage model that predicts the end-effector keypose and then a keypose-conditioned trajectory. PointFlowMatch and DP3 are trained to predict closed-loop trajectories.

\subsection{Detailed PerAct results}
\label{sec:detailed_peract_results}
\begin{table*}[t]
    \centering
    % \vspace{-10pt}
    \begin{adjustbox}{width=\textwidth}
    \tb{@{}l|c|c|c|c|c|c|c|c|c|c@{}}{1.0}{
    & \cellcolor{ImportantColor}Avg. & open & slide & sweep to & meat off & turn & put in & close & drag & stack \\
    & \cellcolor{ImportantColor}Success & drawer & block & dustpan & grill & tap & drawer & jar & stick & blocks \\

    \midrule
    3DDA~\cite{ke20243d} & \cellcolor{ImportantColor}\textbf{81.3} & 89.6$_{\pm 4.1}$ & 97.6$_{\pm 3.2}$ & 84.0$_{\pm 4.4}$ & \textbf{96.8}$_{\pm 1.6}$ & \textbf{99.2}$_{\pm 1.6}$ & \textbf{96.0}$_{\pm 3.6}$ & 96.0$_{\pm 2.5}$ & \textbf{100.0}$_{\pm 0.0}$ & \textbf{68.3}$_{\pm 3.3}$ \\

    \method{} (2-cam) & \cellcolor{ImportantColor}78.4 & 95.2$_{\pm 3.4}$ & \textbf{98.4}$_{\pm 2.2}$ & 93.6$_{\pm 1.8}$ & \textbf{96.8}$_{\pm 3.4}$ & \textbf{99.2}$_{\pm 1.8}$ & 94.4$_{\pm 4.6}$ & 92.0$_{\pm 2.2}$ & 99.2$_{\pm 1.8}$ & 28.0$_{\pm 3.6}$ \\

    \method{} (4-cam) & \cellcolor{ImportantColor}80.3 & \textbf{97.2}$_{\pm 2.5}$ & \textbf{98.4}$_{\pm 2.2}$ & \textbf{99.2}$_{\pm 1.8}$ & \textbf{96.8}$_{\pm 1.8}$ & 97.8$_{\pm 1.8}$ & 95.2$_{\pm 3.0}$ & \textbf{98.4}$_{\pm 2.2}$ & 98.0$_{\pm 2.5}$ & 41.2$_{\pm 3.9}$ \\

    \midrule\midrule
    & & screw & put in & place & put in & sort & push & insert & stack & place \\
    & & bulb & safe & wine & cupboard & shape & buttons & peg & cups & cups \\

    \midrule
    3DDA~\cite{ke20243d} & & \textbf{82.4}$_{\pm 2.0}$ & 97.6$_{\pm 2.0}$  & 93.6$_{\pm 4.8}$ & \textbf{85.6}$_{\pm 4.1}$ & 44.0$_{\pm 4.4}$ & 98.4$_{\pm 2.0}$ & \textbf{65.6}$_{\pm 4.1}$ & 47.2$_{\pm 8.5}$ & 24.0$_{\pm 7.6}$ \\

    \method{} (2-cam) & & 80.0$_{\pm 8.0}$ & \textbf{98.4}$_{\pm 2.2}$  & \textbf{98.4}$_{\pm 2.2}$ & 68.8$_{\pm 7.7}$ & 42.0$_{\pm 4.9}$ & \textbf{99.2}$_{\pm 1.8}$ & 56.4$_{\pm 2.2}$ & 46.4$_{\pm 2.2}$ & 24.0$_{\pm 2.8}$ \\

    \method{} (4-cam) & & 72.4$_{\pm 4.4}$ & 93.6$_{\pm 4.1}$  & 95.4$_{\pm 2.2}$ & 78.6$_{\pm 5.4}$ & \textbf{46.4}$_{\pm 4.1}$ & 97.8$_{\pm 1.8}$ & 59.2$_{\pm 3.0}$ & \textbf{56.0}$_{\pm 2.5}$ & \textbf{24.2}$_{\pm 5.3}$ \\
    \bottomrule
    }
    \end{adjustbox}
    \caption{\textbf{Detailed results on the single-arm PerAct benchmark.} 
    \method{} performs on par with 3DDA on most tasks, even when trained and tested with only two cameras.
    }
    \label{tab:peract}
\end{table*}

We show detailed results on all 18 tasks for 3DDA and \method{} in Table~\ref{tab:peract}. We observe that the average success rate is not fully informative. Although the results in many tasks align, there are statistically significant differences on tasks such as block stacking, putting groceries into cupboard, screwing bulb, sweeping to dustpan and opening drawer. Variance is larger for the two-camera \method{}, as a single mistake may cause severe occlusions across all views, a scenario that is rarer when more cameras are used. Throughout our experimentation, we noticed different sources of large variance:
\begin{itemize}
    \item First, we identified a very large variance in performance on those tasks across checkpoints. For example, for the same model variant, one checkpoint can achieve 80\% success in screwing a bulb, while another can achieve 60\%. The pattern we observed is that the variance across checkpoints on individual tasks is large, but the average performance on all tasks does not vary significantly. Our hypothesis is that 3DDA and \method{} are very low-parameter models: during multi-task training, the network may pick a ``mode", where it adapts to specific tasks more and underfits others. Understanding how low-capacity models select task-specific modes during multi-task training is an important avenue for future investigation.

    \item Second, we tried to fix a checkpoint and evaluate different seeds, which is how we calculate the variance in Table~\ref{tab:peract}. Most tasks display relatively low variance, e.g., they may hit or miss one episode, but this is exacerbated by the small amount of test episode that PerAct offers: with 25 test episodes, a hit/miss contributes 4\% to a task's performance. Interestingly, we find that this variance is not due to the non-deterministic nature of 3DDA or \method{}, but mostly due to the interaction with the simulator, including the motion planner.

    \item To isolate the effect of seed, we run the same checkpoint several times with a fixed seed. We still find that internal states in the RLBench simulator and the planner are not controlled by seeding, thus we still observe variance in many tasks.
\end{itemize}

\subsection{Detailed HiveFormer results}
\label{sec:detailed_hiveformer_results}
\begin{figure*}[t]
    % \vspace{-25pt}
    \imwjpg{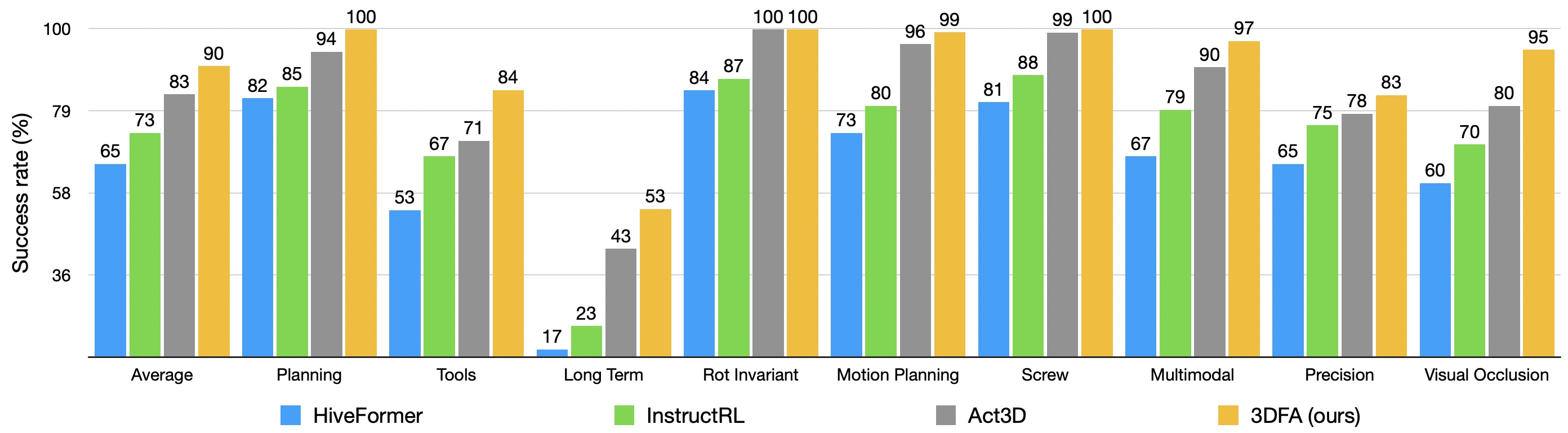}{1.0}
    \caption{\textbf{Detailed evaluation on 74 RLBench tasks.} grouped into 9 categories. \method{} achieves a new state of the art on all groups.}
    \label{fig:hiveformer74}
\end{figure*}
\begin{table*}[t]
    \centering
    % \vspace{-5pt}
    \begin{adjustbox}{width=\textwidth}
    \tb{@{}cc|cc|cc@{}}{1.0}{

        stack cups & 84 & beat the buzz & 100 & lift numbered block & 100 \\

        turn tap & 100 & pick up cup & 100 & wipe desk & 37 \\

        stack blocks & 85 & take shoes out of box & 71 & slide cabinet open and place cup & 20 \\

        close microwave & 100 & close fridge & 100 & close grill & 100 \\

        unplug charger & 99 & open grill & 100 & press switch & 100 \\

        put money in safe & 99 & take money out of safe & 100 & open microwave & 100 \\

        close door & 96 & open door & 100 & open fridge & 70 \\

        open oven & 99 & plug charge in power supply & 60 & slide block to target & 100 \\

        take frame off hanger & 96 & reach and drag & 100 & water plants & 79 \\

        hang frame on hanger & 50 & scoop with spatula & 86 & place hanger on rack & 88 \\

        move hanger & 100 & sweep to dustpan & 100 & take plate off colored dish rack & 100 \\

        screw nail & 26 & toilet seat down & 100 & close laptop lid & 100 \\

        open drawer & 98 & open box & 100 & close drawer & 100 \\

        phone on base & 97 & close box & 100 & reach target & 100 \\

        toilet seat up & 98 & push button & 100 & lamp on & 100 \\

        put books on bookshelf & 99 & lamp off & 100 & pick and lift & 100 \\

        take lid off saucepan & 100 & turn oven on & 100 & change clock & 100 \\

        open window & 100 & open wine bottle & 100 & basketball in hoop & 100 \\

        put rubbish in bin & 100 & meat off grill & 100 & meat on grill & 100 \\

        change channel & 100 & tv on & 100 & tower3 & 100 \\

        push buttons & 100 & stack wine & 100 & take usb out of computer & 100 \\

        insert onto square peg & 80 & play jenga & 100 & take umbrella out of umbrella stand & 100 \\

        insert usb in computer & 78 & straighten rope & 75 & pick and lift small & 100 \\

        put knife on chopping board & 99 & place shape in shape shorter & 40 & take toilet roll off stand & 97 \\

        put umbrella in umbrella stand & 25 & setup checkers & 99 & - & - \\
    }
    \end{adjustbox}
    \caption{\textbf{Detailed results of \method{} on all 74 single-arm tasks.}
    }
    \label{tab:hf_full}
\end{table*}

We show the results of \method{} and baselines on the 9 groups in Figure~\ref{fig:hiveformer74} and the results of \method{} on all 74 tasks in Table~\ref{tab:hf_full}. \method{} is trained to predict trajectories for all tasks and performs very well on most of them. Predicting trajectories allows \method{} to not rely on motion planners or hand-designed collision checking that previous works employ \cite{gervet2023act3d}. While \method{} solves 57 of 74 tasks with over 90\% accuracy, several tasks still exhibit lower success rates. We analyze these failures in detail in Section~\ref{sec:failure}.

\subsection{Detailed real-world results}
\label{sec:detailed_rw_results}

As shown in Table~\ref{tab:realworld}, \method{} excels in performing easy tasks, including \textit{lift ball}, \textit{straighten rope}, \textit{pick up plate}, \textit{stack bowl} and \textit{put marker into bowl}. For more challenging tasks such as \textit{stack blocks}, \textit{open marker}, \textit{close ziploc} and \textit{insert battern}, where objects are smaller or transparent or complex contact dynamics are involved, the performance of our method and baselines decreases,  indicating spatial and dynamic reasoning as a clear avenue for future work.

It is worth mentioning that, when \method{} fails to grasp the related object, it often automatically re-attempts the grasp. The model will re-attempt the task until it is successful or until the object moves out of the bounding box. This is a result of training for closed-loop trajectory prediction and is not observed when we optimize for keypose prediction.

\subsection{Failure cases}
\label{sec:failure}
We analyze the failure modes of \method{} in both the simulation and the real world. We also discuss the failure modes of our baselines in the real world.

\textbf{On PerAct2}, we notice that most of the errors come from motion planning. For example, in the ``handover item" tasks, the RRT-predicted trajectories may not be collision-free when the object is placed near the receiving arm. The task with the lowest performance is ``straighten rope". RRT predicts linear segments that do not respect the rope limits. Interestingly, we achieve a much higher success rate for the same task in the real world, where we predict trajectories directly. We did not train a trajectory model on PerAct2 because we found the interaction with the simulator to be particularly slow, making the evaluation of a trajectory prediction model impractical.

To validate the effect of the RRT planner in this setup, we replayed the ground-truth keyposes for the test episodes and use the planner to move between them. We get an average performance of 85.8\%, which is only marginally better than our model's. ``Pick up plate" and ``straighten rope" are still the tasks with most errors, achieving 59\% and and 39\% respectively.

\textbf{On PerAct}, we observe that the model struggles with understanding (sometime unseen) variations at test time. For example, it may place the lid on the wrong jar in ``close jar" or stack blocks of wrong colors. We verify that on the HiveFormer setup, where there are no variations, \method{} achieves very high success rates on stacking blocks and cups (Table~\ref{tab:hf_full}). This could be mitigated by integrating more powerful representations from a VLM, rather than CLIP. Another source of failure is lower precision than needed in some tasks, such as in ``put a block in the shape sorter" and ``insert onto peg". \method{} featurizes all visual observations and subsamples the feature cloud, resulting in a sparser representation. Reducing the number of points is necessary to manage computational resources. This could be prevented using some VLM guidance to focus around regions of interest for the next action. Lastly, when using two cameras only, occlusions can become an issue.

\textbf{On the 74 HiveFormer tasks}, high-precision requirements and occlusions are the main issues. \method{} achieves its lowest performance on the ``long-term" task group. However, this is due to heavy occlusions caused by using the front and wrist cameras only. Specifically, for ``slide cabinet open and place cup", the cup is most of the times hidden from the front camera, as it is occluded by the cabinet. \method{} always opens the cabinet but then struggles to find the cup. To verify, we trained \method{} using the overhead camera for this task and achieved 74\%.  Another example of heavy occlusion is ``open fridge". Some examples of high-precision tasks that \method{} struggles are ``hang frame on hanger", ``place shape in shape shorter" and ``put umbrella in umbrella stand".

\textbf{In the real world} experiments, most failure cases of \method{} in simple tasks are caused by last-centimeter errors that move the objects towards an undesired direction. Examples include slipping the ball or not raising the other edge of the plate high enough. For hard tasks, the success rate is much lower; we think this is caused by occlusion from the robot or objects in the ``handover block", ``open marker" and ``insert battery" tasks. We also discovered a 2cm error in the robot's forward kinematics, which significantly reduces the performance on the aforementioned tasks that require precise operation. This error is exacerbated when the arm leans forward too much. We show videos of different failure cases for our model on our website.

\textbf{For baselines}, we found that $\pi_0$~\cite{black2024pi_0} requires the relevant object to always remain visible in the wrist camera view, otherwise it cannot finish the grasping part; this problem is obvious in ``handover block", ``open marker" and ``insert battery"  when the wrist camera view is blocked by the object in hand or the robot itself. Most of $\pi_0$ failures stem from difficulty in understanding 3D relationships between objects. For example, in the two stacking tasks, the object cannot be placed directly above the other, and in the "insert marker into the cup" task, the model cannot find the correct position to drop the marker. On the other hand, iDP3~\cite{ze2024generalizable} has a high success rate in finding the 3D relationship between objects, while most of its failure cases are caused by failed grasps.

\end{document}